\documentclass[lettersize,journal]{IEEEtran}
\usepackage{amsmath,amsfonts}
\usepackage{algorithmic}
\usepackage{algorithm}
\usepackage{array}
\usepackage[caption=false,font=normalsize,labelfont=sf,textfont=sf]{subfig}
\usepackage{textcomp}
\usepackage{stfloats}
\usepackage{url}
\usepackage{verbatim}
\usepackage{graphicx}
\usepackage{cite}
\usepackage{hyperref}       
\usepackage{url}            

\usepackage{algorithm}
\usepackage{algorithmic}

\usepackage{bbding}
\usepackage{utfsym}
\usepackage{amsmath}
\usepackage{color}
\usepackage{xcolor}         
\usepackage{colortbl}
\definecolor{tabgray1}{gray}{.9}
\definecolor{tabgray2}{gray}{.84}
\usepackage{multirow} 
\usepackage{makecell}
\usepackage{pifont}
\definecolor{green}{rgb}{0.0,1.0,0.0}
\newcommand{\green}[1]{{\color{green} #1}}
\definecolor{redd}{rgb}{1.0,0.0,0.0}
\newcommand{\redd}[1]{{\color{redd} #1}}
\newcommand{\cmark}{\green{\ding{51}}}
\newcommand{\xmark}{\redd{\ding{55}}}
\usepackage{bm}
\definecolor{mygray}{gray}{.7}
\definecolor{mypink}{rgb}{.99,.91,.95}
\definecolor{mycyan}{cmyk}{.3,0,0,0} 
\usepackage{calligra}
\usepackage{color}
\usepackage{amssymb}
\definecolor{BlueColor}{rgb}{0.0, 0.0, 1.0}
\definecolor{RedColor}{rgb}{1.0, 0.0, 0.0}
\definecolor{GreenColor}{rgb}{0.3, 0.6, 0.4}

\usepackage{lineno}
\usepackage{booktabs}
\usepackage{xspace} 
\newcommand{\eg}{\emph{e.g.}\xspace}
\newcommand{\ie}{\emph{i.e.}\xspace}

\usepackage{newfloat}
\usepackage{listings}
\usepackage{hyperref}

\usepackage{bibentry}

\begin{document}

\title{$\text{H}^3$Former: Hypergraph-based Semantic-Aware Aggregation \\ via Hyperbolic Hierarchical Contrastive Loss \\ for Fine-Grained Visual Classification}
\author{Yongji Zhang\textsuperscript{†}, Siqi Li\textsuperscript{†}, Kuiyang Huang, Yue Gao, \IEEEmembership{Senior Member, IEEE} and Yu Jiang*
\thanks{This work was supported by Brain Science and Brain-like Intelligence Technology National Science and Technology Major Project (2025ZD0217300), National Natural Science Foundation of China (Nos. U25A20532 and 62501358), the Beijing Natural Science Foundation under Grant No. L242167, the Open Project Program of State Key Laboratory of Virtual Reality Technology and Systems, Beihang University (No.VRLAB2025A01), and the National Natural Science Foundation of China under Grant 62072211. (\textit{Corresponding author: Yu Jiang.})}
\thanks{† These authors contributed equally to this work.}
\thanks{Yongji Zhang and Yu Jiang are with the College of Computer Science and Technology, Jilin University, Changchun 130012, China (zhangyongji1998@gmail.com;~jiangyu2011@jlu.edu.cn).}
\thanks{Kuiyang Huang is with the College of Software, Jilin University, Changchun 130012, China (huangky24@mails.jlu.edu.cn).}
\thanks{Siqi Li and Yue Gao are with BNRist, THUIBCS, BLBCI, School of Software, Tsinghua University, Beijing 100084, China (lisiqi19971013@gmail.com; kevin.gaoy@gmail.com).}}

\maketitle

\begin{abstract}

Fine-Grained Visual Classification (FGVC) remains a challenging task due to subtle inter-class differences and large intra-class variations. Existing approaches typically rely on feature-selection mechanisms or region-proposal strategies to localize discriminative regions for semantic analysis. However, these methods often fail to capture discriminative cues comprehensively while introducing substantial category-agnostic redundancy. To address these limitations, we propose $\text{H}^3$Former, a novel token-to-region framework that leverages high-order semantic relations to aggregate local fine-grained representations with structured region-level modeling. Specifically, we propose the Semantic-Aware Aggregation Module (SAAM), which exploits multi-scale contextual cues to dynamically construct a weighted hypergraph among tokens. By applying hypergraph convolution, SAAM captures high-order semantic dependencies and progressively aggregates token features into compact region-level representations. Furthermore, we introduce the Hyperbolic Hierarchical Contrastive Loss (HHCL), which enforces hierarchical semantic constraints in a non-Euclidean embedding space. The HHCL enhances inter-class separability and intra-class consistency while preserving the intrinsic hierarchical relationships among fine-grained categories. Comprehensive experiments conducted on four standard FGVC benchmarks validate the superiority of our $\text{H}^3$Former framework. Code is available at \href{https://github.com/xiaozhangfangyang/H3Former}{https://github.com/xiaozhangfangyang/H3Former}.

\end{abstract}

\section{Introduction}
\begin{figure}[t] 
\centering 
\includegraphics[width=1.0\linewidth]{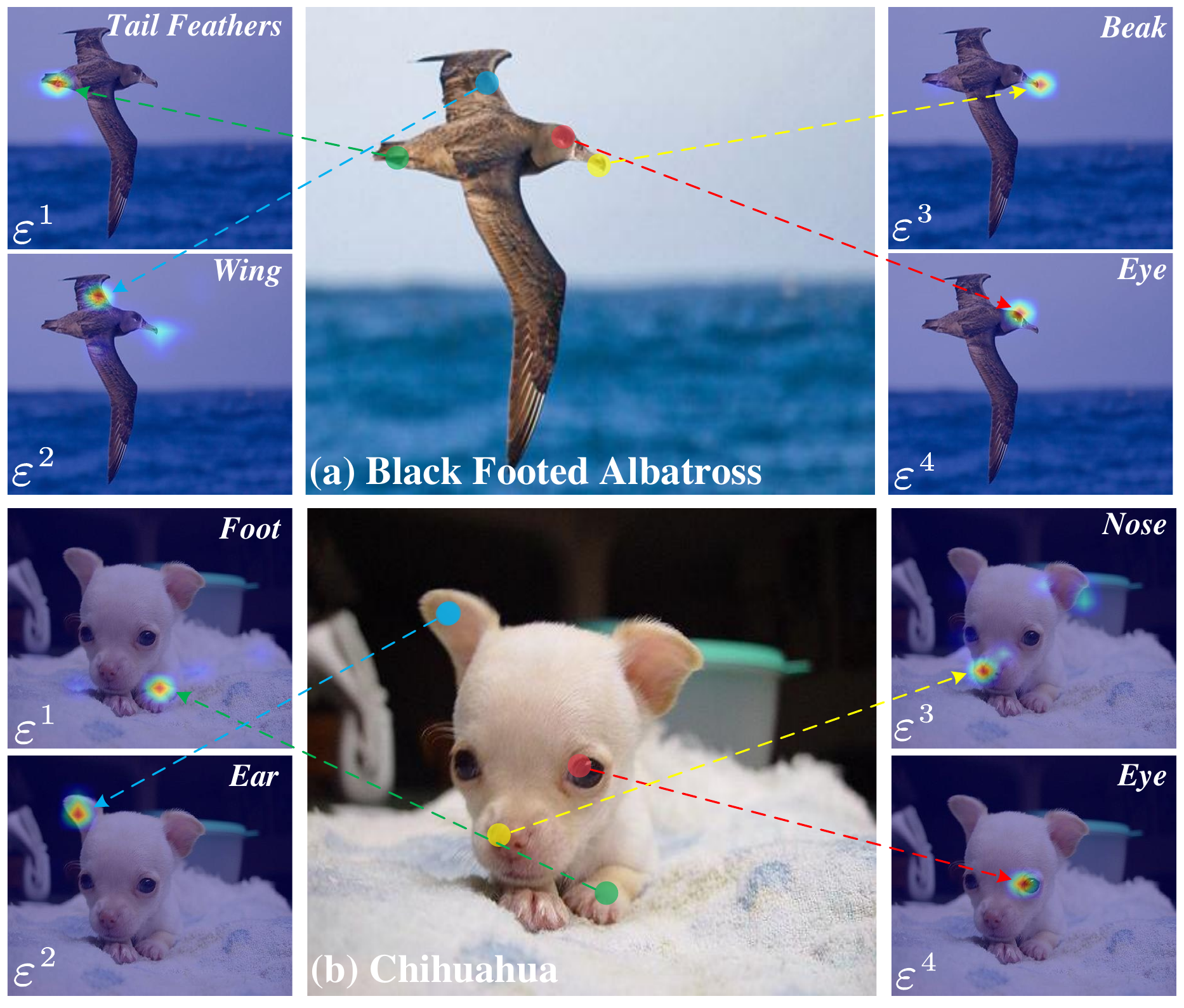} 
\vspace{-0.5cm}
\caption{\textbf{Hyperedges ($\mathcal{E}^1$–$\mathcal{E}^4$) of hypergraph $\mathcal{H}=(\mathcal{V},\mathcal{E})$ generated by our $\text{H}^3$Former.} Distinct hyperedges correspond to meaningful semantic regions,~\eg, tail feathers, wing, beak, and eye. The learned hypergraphs automatically highlight key discriminative parts without any part-level supervision. $\text{H}^3$Former adaptively constructs coherent semantic regions through its hypergraph construction mechanism, bridging local token cues and global structural representation for FGVC.}
\vspace{-0.4cm}
\label{fig:new} 
\end{figure}

Fine-Grained Visual Classification (FGVC) aims to distinguish subordinate categories within a general class, \eg, distinguishing \textit{the Black-footed Albatross} in Fig.~\ref{fig:new} (a) from its close relative, \textit{the Sooty Albatross}. Unlike generic object classification, FGVC heavily relies on capturing subtle visual differences typically localized in structural or textural cues. This task requires models to possess fine-grained spatial sensitivity and robust detail modeling capabilities. Additionally, challenges such as subtle inter-class differences, significant intra-class variations, limited annotated data, and complex background clutter significantly complicate the task. Thus, developing robust models capable of identifying discriminative patterns is essential for advancing FGVC performance~\cite{han2022survey,wei2021survey2}.

\begin{figure*}[t] 
\centering 

\includegraphics[width=1.0\linewidth]{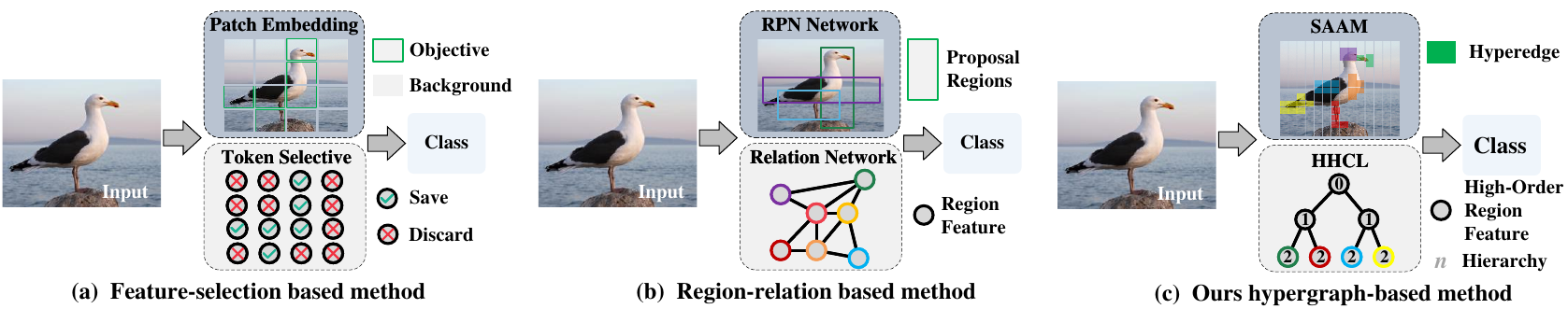} 
\vspace{-0.6cm}
\caption{\textbf{Illustration of different FGVC paradigms.}
(a) Feature-selection based methods perform token filtering in the feature space to retain features most relevant to fine-grained recognition, but overlook coherent semantic structure.
(b) Region-relation based methods learn pairwise dependencies among predefined regions, typically obtained from RPNs, which may introduce redundant and category-agnostic information.
(c) Our proposed $\text{H}^3$Former organizes discrete tokens into structured semantic regions via a hypergraph formulation, where each hyperedge adaptively aggregates related tokens. Furthermore, the proposed HHCL imposes hierarchical constraints to enhance the discriminability and consistency of these regions.}
\vspace{-0.3cm}
\label{fig:01} 
\end{figure*}

Recent advancements in Vision Transformers (ViTs) have significantly advanced feature-selection based FGVC methods~\cite{zhao2017survey1, khan2022transformers, accvit, mpt, wang2021ffvt}, \eg, TransFG~\cite{he2022transfg} identifies the most informative tokens by aggregating attention maps across multiple transformer layers, whereas IELT~\cite{xu2023ielt} fuses multi-head attention weights with feature cues to guide the localization of key regions. As illustrated in Fig.~\ref{fig:01} (a), these methods exploit the self-attention mechanism of ViTs to preserve tokens corresponding to discriminative parts for FGVC. However, due to the inherently local and fragmentary semantics represented by individual tokens, feature-selection based approaches often isolate discrete tokens and fail to capture the discriminative regions comprehensively.

An alternative research direction is region-proposal based methods, which generate candidate regions through category-agnostic or category-aware Region Proposal Networks (RPNs), \eg, LGTF~\cite{lgtf} employs a region selection gate for filtering after RPNs. Although explicit region modeling enhances feature discriminability, it also introduces substantial redundant background information, which distracts the model from truly discriminative cues and consequently reduces efficiency and recognition performance. As shown in Fig.~\ref{fig:01} (b), recent methods, \eg, SR-GNN~\cite{srgnn} and I2-HOFI~\cite{i2o}, incorporate a graph-based region-relation network to align and refine proposal features. However, conventional graph convolutional networks are inherently limited to pairwise aggregation and thus fail to effectively capture higher-order dependencies among different regions across the entire image~\cite{buhyper8,buhyper9,hgformer}.

This observation motivates a unified approach that combines token- and region-level modeling via a semantic-aware mechanism, which adaptively aggregates informative tokens into coherent discriminative regions. Thus, we introduce a \textbf{H}ypergraph-based Semantic-Aware Aggregation Module (SAAM) via \textbf{H}yperbolic \textbf{H}ierarchical Contrastive Loss (HHCL) for FGVC, referred to as \textbf{$\text{H}^3$Former}. It provides a token-to-region representation framework that bridges the structural gap between token-level modeling and region-level representation in FGVC.

As illustrated in Fig.~\ref{fig:01} (c), our $\text{H}^3$Former is organized in a hypergraph manner, where each hyperedge adaptively connects semantically related tokens to form coherent regions. Specifically, the proposed SAAM leverages multi-scale contextual cues to initialize learnable hyperedge prototype vectors. By measuring the similarity between visual tokens and these prototypes, the model dynamically constructs semantically enriched hyperedges. Each hyperedge connects all visual tokens through learnable participation weights, thereby adaptively modeling high-order semantic relations within visual features. 

Moreover, we propose the HHCL, which works synergistically with SAAM to constrain the semantic representations of the obtained region features, thereby enhancing feature discriminability. As illustrated in Fig.~\ref{fig:01} (c), the semantic region features derived from SAAM are treated as leaf nodes and are progressively merged to construct multi-level hierarchical representations. The HHCL applies contrastive constraints in both Euclidean and hyperbolic spaces across these hierarchical levels to maximize intra-class similarity and inter-class separability. At the same time, a parent–child consistency constraint is imposed to maintain structural coherence within the hierarchy. Through this joint supervision, the model learns to achieve a semantically smooth transition from local details to global concepts. This geometry-aware objective complements the semantic aggregation performed by SAAM, enabling $\text{H}^3$Former to learn well-organized, hierarchy-aware, and highly discriminative representations even under subtle visual variations.

Overall, $\text{H}^3$Former unifies semantic-aware token aggregation and geometry-aware representation learning within a single framework, where hypergraph-guided region construction and hierarchical contrastive optimization jointly enhance generalization and discriminative capability in FGVC.

As shown in Fig.~\ref{fig:new}, we visualize the learned hypergraphs on two representative FGVC datasets. Each hyperedge in $\text{H}^3$Former adaptively corresponds to a semantic region formed by aggregating correlated tokens with similar semantics. On \textit{the Black-footed Albatross} from CUB-200-2011~\cite{wah2011cub200} dataset, different hyperedges ($\mathcal{E}^1$–$\mathcal{E}^4$) distinctly focus on body parts,~\eg, tail feathers, wings, beak, and eyes, reflecting the model's ability to capture semantically correlated components. Similarly, on \textit{the Chihuahua} from Stanford-Dogs~\cite{dogs} dataset, our model automatically localizes discriminative regions including the ear, nose, and foot, even without explicit part annotations. These visualizations provide strong empirical evidence that the proposed hypergraph construction mechanism successfully captures region-level semantic structures, effectively bridging the gap between local appearance cues and holistic object understanding.

Our main contributions are summarized as follows:

\begin{itemize}

\item We introduce $\text{H}^3$Former, an FGVC-oriented token-to-region representation framework that adaptively structures token-level cues into semantically coherent region-level representations.

\item We propose the Semantic-Aware Aggregation Module (SAAM), which employs hypergraph to capture high-order relations among tokens and progressively form semantically coherent region representations.

\item We design the Hyperbolic Hierarchical Contrastive Loss (HHCL) to enhance hierarchical semantic representations and improve class discriminability in both Euclidean and hyperbolic spaces.

\item Extensive experiments on four widely used FGVC benchmarks, including CUB-200-2011, NA-Birds, Stanford-Dogs, and OXford Flowers-101 validate the effectiveness of H$^3$former.
\end{itemize}

\section{Related Work}

\subsection{Feature-selection based Methods}
Vision Transformers (ViTs)~\cite{vaswani2017attention} adapt Transformers to vision by representing images as tokens and modeling long-range dependencies via self-attention. While effective for global context modeling, their weak spatial inductive bias limits the performance of Fine-Grained Visual Classification (FGVC)~\cite{han2022survey,zhao2017survey1,wei2021survey2}. 

To address this limitation, various approaches have been proposed to enhance token selection and representation. TransFG~\cite{he2022transfg} employs overlapping patches and attention-based selection to highlight informative regions. RAMS~\cite{hu2021rams} locates object-centered sub-images based on attention heatmaps and re-feeds them to suppress background noise. IELT~\cite{xu2023ielt} fuses multi-head attention weights with token features to better infer regional importance. FFVT~\cite{wang2021ffvt} incorporates low-level feature cues to refine token response maps for part-aware selection. More recent works explore richer semantic modeling. MP-FGVC~\cite{mpFGVC} introduces vision-language prompts to guide token discrimination across modalities, aligning visual and textual semantics for cross-modal reasoning. ACC-ViT~\cite{accvit} integrates attention patch mixing, region filtering, and multi-level token fusion to address background noise and token complementarity. MpT-Trans~\cite{mpt} replaces the class token with multiple part tokens via a Part-wise Shift Learning module, further enhanced by a dual contrastive loss that improves both feature diversity and fine-grained discrimination. Recent efforts further introduce semantic priors or cross-modal cues for more robust fine-grained representation learning.~\eg, LearnMat~\cite{learnmat} leverages vision-language semantic distillation and gradient-based signals to suppress irrelevant patterns and highlight subtle discriminative regions under a self-supervised FGVR setting. In addition, attention-guided pyramidal feature learning~\cite{learning} explores multi-scale local cues for few-shot fine-grained recognition, while recent multimodal knowledge transfer methods~\cite{tang2025connecting} further investigate how large multimodal models can provide semantic guidance for few-shot representation learning.

Unlike previous token-based approaches, our method employs hypergraph-guided aggregation to capture high-order relationships among multiple tokens, thereby forming semantically coherent regions that exhibit stronger discriminability.

\subsection{Region-proposal based Methods}

Region-based modeling is another core strategy for FGVC, focusing on discovering and leveraging discriminative object parts to distinguish between visually similar subcategories. Early approaches heavily rely on strong supervision, such as bounding boxes or part annotations~\cite{a1,a2,a3,a4,mcl,liu2021cpcnn}, which are used to align semantic regions across images. For instance, Mask-CNN~\cite{mask} utilizes annotated part masks to guide the selection of convolutional descriptors, achieving compact and effective region-level aggregation. Similarly, multi-branch architectures have been proposed~\cite{a3,a4} to process individual parts separately before fusing them for final classification. To alleviate the annotation burden, recent works have explored weakly or self-supervised part discovery. These methods typically use class activation maps or attention mechanisms to localize salient regions without explicit labels. PART~\cite{part} introduces a unified framework that combines gradient-based part localization with relational Transformers, enabling semantic interaction between global and part-level features without additional inference overhead. Other approaches, such as PMRC~\cite{pmrc} use graph reasoning over selected regions to capture implicit structural relations among parts in a weakly supervised manner. MPSA~\cite{mpsa} proposes multi-granularity part sampling attention to extract discriminative parts with different scales and shapes, reducing the limitation of rectangular region extraction. SPA~\cite{spa} further explores semantic-part alignment by associating part embeddings with fine-grained semantic representations and regularizing their relation through optimal transport. For retrieval-oriented fine-grained recognition, DAHNet~\cite{jiang2024global} jointly exploits global and local discriminative cues to improve large-scale fine-grained image retrieval.

Beyond region proposal and part discovery, recent works have explored modeling the structural relationships among regions to capture fine-grained object semantics better. For instance, SR-GNN~\cite{srgnn} introduces a graph-based framework that integrates relation-aware feature transformation and context-aware attention to aggregate discriminative cues from semantically relevant regions. Similarly, I2-HOFI~\cite{i2o} constructs both inter- and intra-region graphs to capture structural hierarchies: inter-region graphs encode long-range contextual dependencies across distinct parts, while intra-region graphs focus on fine-grained local relationships within each region. SIM-OFE~\cite{sim} also emphasizes internal object structure modeling by mining the distribution and contextual relations of critical regions and enhancing object-aware features for FGVC. These complementary graphs are jointly optimized via message passing to improve region-level feature discrimination.

Despite their effectiveness, existing region-structure methods mainly rely on attention, proposal sampling, or graph-based pairwise modeling, which limits their ability to aggregate multiple correlated tokens into high-order semantic structures. In contrast, our hypergraph formulation adaptively connects semantically related tokens within each hyperedge, enabling high-order aggregation from token-level cues to region-level representations.

\subsection{Hypergraph Networks}

Hypergraphs offer a natural way to model high-order relationships, as each hyperedge can simultaneously connect multiple nodes, making them well-suited for capturing group-wise semantic interactions beyond pairwise graphs. Hypergraph Neural Networks (HGNNs) extend this representation with a vertex–hyperedge–vertex message passing mechanism, enabling richer structural reasoning than conventional GNNs~\cite{hyper2,buhyper2,buhyper3,buhyper4}.

While HGNNs have shown strong performance in non-visual domains such as social networks and bioinformatics, their application to visual recognition is still relatively nascent. Recent efforts have explored injecting hypergraph structures into convolutional~\cite{hyperyolo} and Transformer-based frameworks~\cite{visionhgnn, hgformer} to enhance long-range dependency modeling. For example, Hyper-YOLO~\cite{hyperyolo} embeds hypergraph computation into the detection neck for cross-level semantic fusion, and Vision HGNN~\cite{visionhgnn} replaces standard Transformer modules with hypergraph convolutions.

In contrast, we propose a dynamic hypergraph formulation that adaptively aggregates semantically related tokens into coherent regions based on high-order dependencies. This semantic-aware design bridges the gap between token- and region-level representations, enabling structured and fine-grained semantic organization. Furthermore, the proposed HHCL is intrinsically coupled with the hypergraph formulation, operating on hierarchical region representations to encode structural dependencies in two spaces.

\begin{figure*}[ht] 
\centering 
\includegraphics[width=1.0\linewidth]{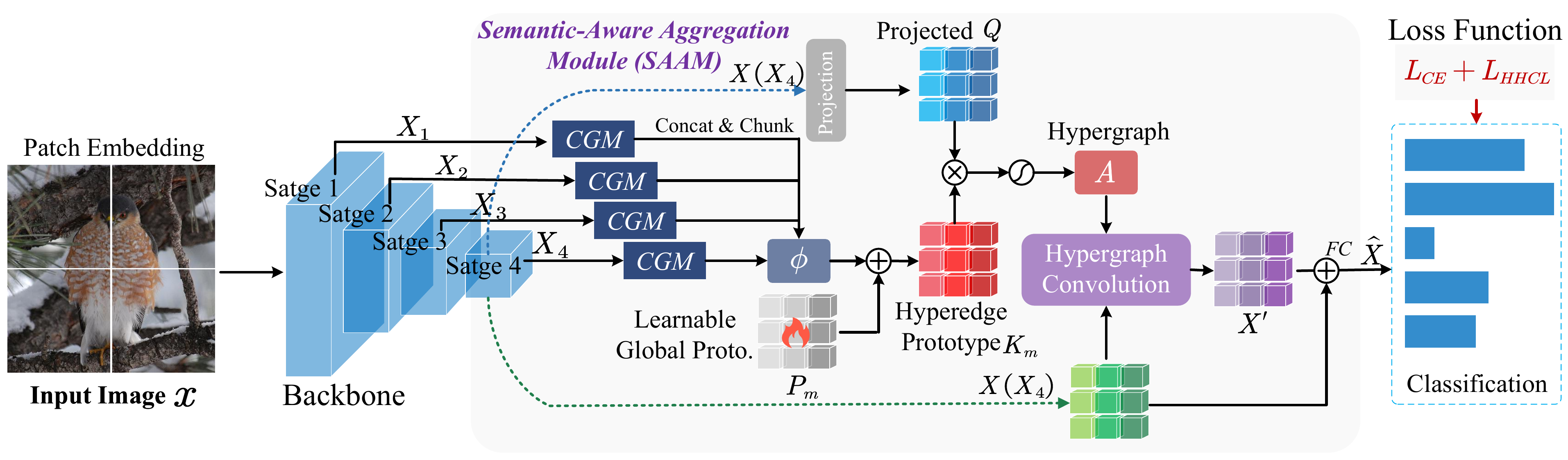} 
\vspace{-0.4cm}

\caption{\textbf{Overview of the proposed $\text{H}^3$Former framework.}
The Semantic-Aware Aggregation Module (SAAM) constructs a weighted hypergraph to capture high-order semantic relations and progressively aggregates tokens into semantically coherent regions. Meanwhile, the Hyperbolic Hierarchical Contrastive Loss (HHCL) operates on the resulting hierarchical region representations to enforce fine-grained category separation and structural consistency in two spaces, yielding more discriminative representations.}
\vspace{-0.2cm}
\label{fig:02} 
\end{figure*}

\section{Method}

In this section, we detail the proposed $\text{H}^3$Former.
We first describe the overall architecture in Sec.~\ref{sec:3.1}, followed by the Semantic-Aware Aggregation Module (SAAM) in Sec.~\ref{sec:3.2}, which constructs semantic regions through hypergraph-based token aggregation. Then,
we introduces the Hyperbolic Hierarchical Contrastive Loss (HHCL) in Sec.~\ref{sec:3.3}, which operates on hierarchical region representations to enforce structured semantic consistency in two spaces.

\subsection{Overall Architecture}
\label{sec:3.1}
The overall framework of $\text{H}^3$Former is illustrated in Fig.~\ref{fig:02}. 
Given an input image $\mathcal{X}$, we adopt a Swin Transformer backbone to extract multi-scale features from four hierarchical stages, denoted as $\{X_1, X_2, X_3, X_4\}$, where each $X_s \in \mathbb{R}^{N_s \times C_s}$ represents $N_s$ tokens with channel $C_s$ at stage $s$.

After feature extraction, we propose the SAAM to bridge the semantic gap between token- and region-level features. As shown in Fig.~\ref{fig:02}, SAAM integrates multi-scale contextual cues through the Context Generation Module (CGM), which produces a set of hyperedge prototypes ${K_m}$. These prototypes interact with the projected token features $Q$ to compute feature similarities, thereby constructing a dynamic weighted hypergraph $\mathcal{H}$ that adaptively captures high-order semantic dependencies among tokens.

Subsequently, message passing is performed through hypergraph convolution, enabling mutual refinement between tokens and semantic regions to produce the refined features $X'$.
A residual connection is then applied to combine $X'$ with the final-stage features, followed by global average pooling and a fully connected classifier to obtain the final prediction $\hat{X}$.

To further enhance inter-class separability and preserve intra-class consistency, we introduce the HHCL, which complements the semantic aggregation performed by SAAM. During HHCL computation, the features of different hyperedges in $\mathcal{H}$ are regarded as leaf nodes and are hierarchically merged to form multi-level region representations. In this hierarchy, lower-level nodes preserve finer-grained local semantics, while higher-level nodes progressively encode more abstract semantic concepts, enabling HHCL to impose cross-level hierarchical supervision rather than only constraining the top-level representation. By performing contrastive learning in both Euclidean and hyperbolic spaces and incorporating a parent–child consistency constraint, HHCL enforces smooth semantic transitions from local to global concepts. Unlike heavy hyperbolic feature transformation pipelines, the proposed HHCL is introduced only as an auxiliary training objective on the high-order semantic regions produced by SAAM.

The overall training objective combines the standard cross-entropy loss $\mathcal{L}_{\text{CE}}$ with the proposed contrastive regularization $\mathcal{L}_{\text{HHCL}}$:
\begin{equation}
\mathcal{L}_{\text{total}} = \mathcal{L}_{\text{CE}} + \alpha \mathcal{L}_{\text{HHCL}},
\end{equation}
where $\alpha$ is a balancing coefficient.

\begin{figure}[t]
\centering 
\includegraphics[width=1\linewidth]{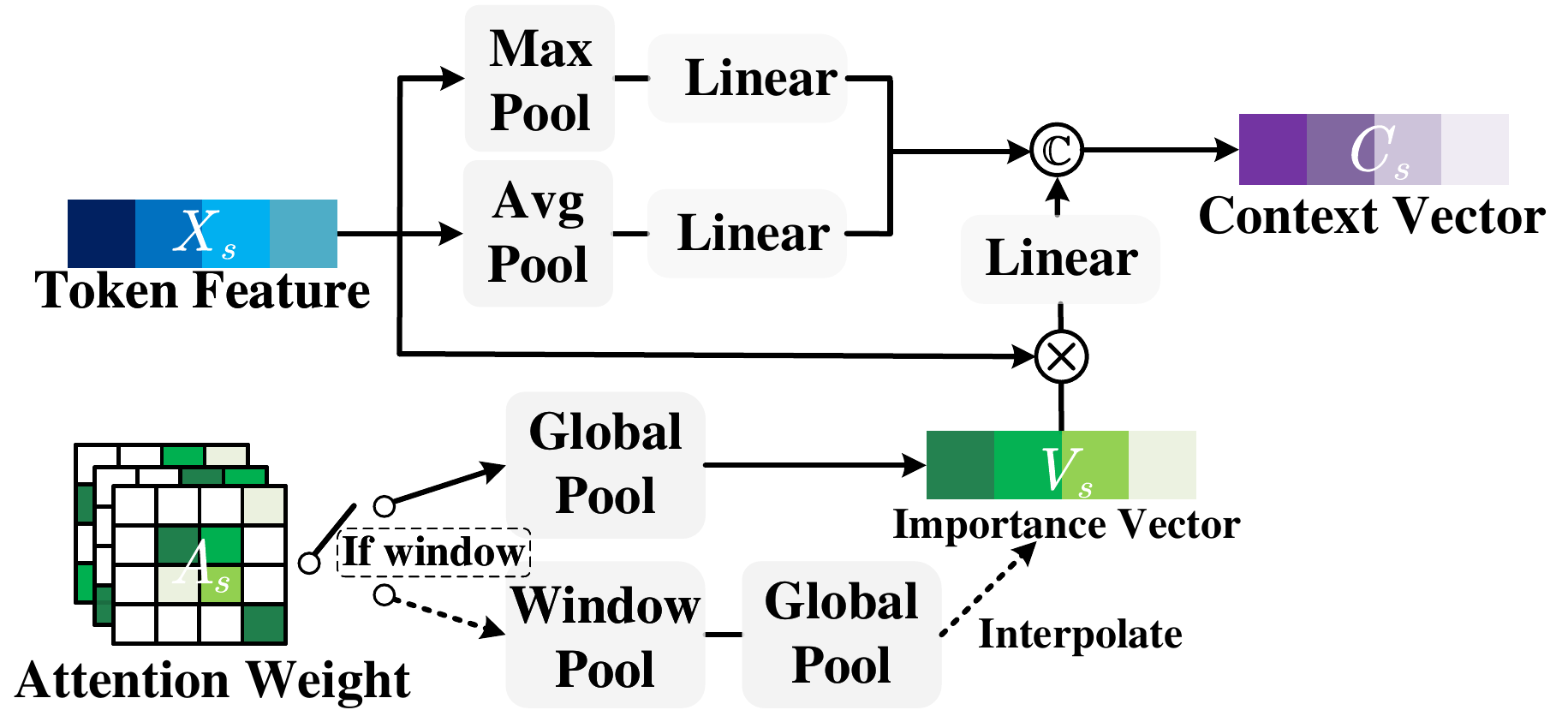} 

\caption{\textbf{The architecture of the Context Generation Module (CGM).} The CGM utilizes the token features and attention maps from each stage to generate corresponding context vectors that encode multi-scale contextual information.
When window-based attention is used, the attention maps are processed along the dashed path to produce the importance vector, which reflects the relative significance of tokens within each window.}
\vspace{-0.2cm}
\label{fig:07} 
\end{figure}

\subsection{Semantic-Aware Aggregation Module (SAAM)}
\label{sec:3.2}

\textbf{Multi-Scale Context Extraction.}
As illustrated in Fig.~\ref{fig:07}, given the stage-wise token features $X_s$ extracted from the backbone, we employ the CGM to obtain the corresponding context vector $C_s$. Specifically, the CGM extracts three complementary types of contextual representations from $X_s$.
The average-pooled context and max-pooled context are obtained by applying average pooling and max pooling to $X_s$, followed by linear projections. For the attention-weighted context, the attention map $A_s$ is first used to estimate a token importance vector $V_s$ (for window-based attention, the attention map is averaged across heads and windows, then interpolated to the full token length). This attention-guided weighting is particularly suitable for FGVC, where subtle discriminative regions are often localized and sparsely distributed. This importance vector $V_s$ is then used to perform weighted aggregation over $X_s$, and the aggregated feature is linearly projected to produce a globally aware attention-weighted context representation.

All context vectors are projected to a unified dimension $C$ and concatenated across stages, resulting in a compact multi-scale context representation:
\begin{equation}
\mathbf{F} = \left\{ f^{\text{avg}}_s, f^{\text{max}}_s, f^{\text{attn}}_s \right\}_{s=1}^S \in \mathbb{R}^{3S \times C},
\end{equation}
where $S$ denotes the number of backbone stages.

\textbf{Semantic Prototype and Hyperedge Generation.} To serve as high-order semantic anchors, the context tensor $\mathbf{F}$ is divided into $M$ channel-wise groups, each representing a semantic subspace that captures specific contextual cues (e.g., texture, color, or part-specific attributes). 
This grouping strategy ensures that the model learns multiple complementary semantic perspectives rather than a single holistic representation. 

Each group $\mathbf{F}_{(m)}$ is first transformed through a shared projection network $\phi(\cdot)$ that performs a lightweight non-linear embedding, aligning all groups into a common latent space of dimension $d_k$. 
Subsequently, we introduce a set of learnable prototype vectors $\{P_m\}_{m=1}^{M}$, where each $P_m \in \mathbb{R}^{d_k}$ acts as a semantic anchor that adaptively represents the centroid of a latent semantic cluster. 
The final semantic prototype of each hyperedge is computed as:
\begin{equation}
K_m = \phi(\mathbf{F}_{(m)}) + P_m, \quad m=1,\dots,M,
\end{equation}
where $\phi(\cdot)$ is shared across groups to encourage semantic consistency while $P_m$ allows flexibility for data-driven adaptation.

Intuitively, these prototypes can be regarded as \textit{semantic attractors} that dynamically summarize contextual patterns across scales. 
Unlike static region templates or fixed part priors, our learnable prototypes evolve jointly with network optimization, enabling adaptive refinement based on dataset-specific distributions. As a result, each prototype $K_m$ defines the centroid of a hyperedge, connecting multiple semantically correlated tokens during the subsequent hypergraph construction.

\textbf{Hypergraph Construction.}

Let $X=X_4 \in \mathbb{R}^{N \times C}$ denote the final-stage token features, which are used as query nodes for computing affinities between tokens and semantic prototypes. We emphasize that SAAM does not rely solely on final-stage information: the hyperedge prototypes $\{K_m\}_{m=1}^{M}$ are generated from the multi-stage context representation $\mathbf{F}$ derived from $\{X_s\}_{s=1}^{S}$ by CGM. Therefore, the association between tokens and prototypes is guided by both shallow fine-grained cues and high-level semantic contexts.

To measure the semantic association between the query tokens and the prototypes $\{K_m\}_{m=1}^M$, we first project $X$ into a query space through a learnable linear transformation $W_q$:
\begin{equation}
Q = \mathbf{X} W_q \in \mathbb{R}^{N \times d_k}.
\end{equation}
This projection aligns the feature dimension with the semantic prototype space, allowing meaningful affinity computation.

We then compute the token–prototype similarity to derive a participation (incidence) matrix $A \in \mathbb{R}^{N \times M}$:
\begin{equation}
A_{i,m} = \frac{\exp(Q_i^\top K_m / \sqrt{d_k})}
{\sum_{m'=1}^{M} \exp(Q_i^\top K_{m'} / \sqrt{d_k})}.
\end{equation}
Each element $A_{i,m}$ quantifies how strongly token $i$ contributes to semantic region $m$. 
Unlike traditional graph adjacency matrices that encode pairwise relations, the participation matrix $A$ naturally defines many-to-many associations between tokens and hyperedges. 
This design allows a single token to simultaneously participate in multiple hyperedges with different degrees of confidence, forming a hypergraph that captures overlapping and complementary semantic patterns.

From a geometric perspective, theassignment process effectively learns a high-order incidence structure where each hyperedge aggregates semantically coherent tokens distributed across spatially distant regions. 
Such flexibility enables the model to adaptively discover meaningful part–whole compositions without explicit supervision or pre-defined region proposals. Consequently, the constructed hypergraph $\mathcal{H}$ provides a unified representation that connects token-level details with region-level semantics, laying the foundation for high-order message passing in the subsequent stage.

\begin{figure}[t]
\centering 
\includegraphics[width=1\linewidth]{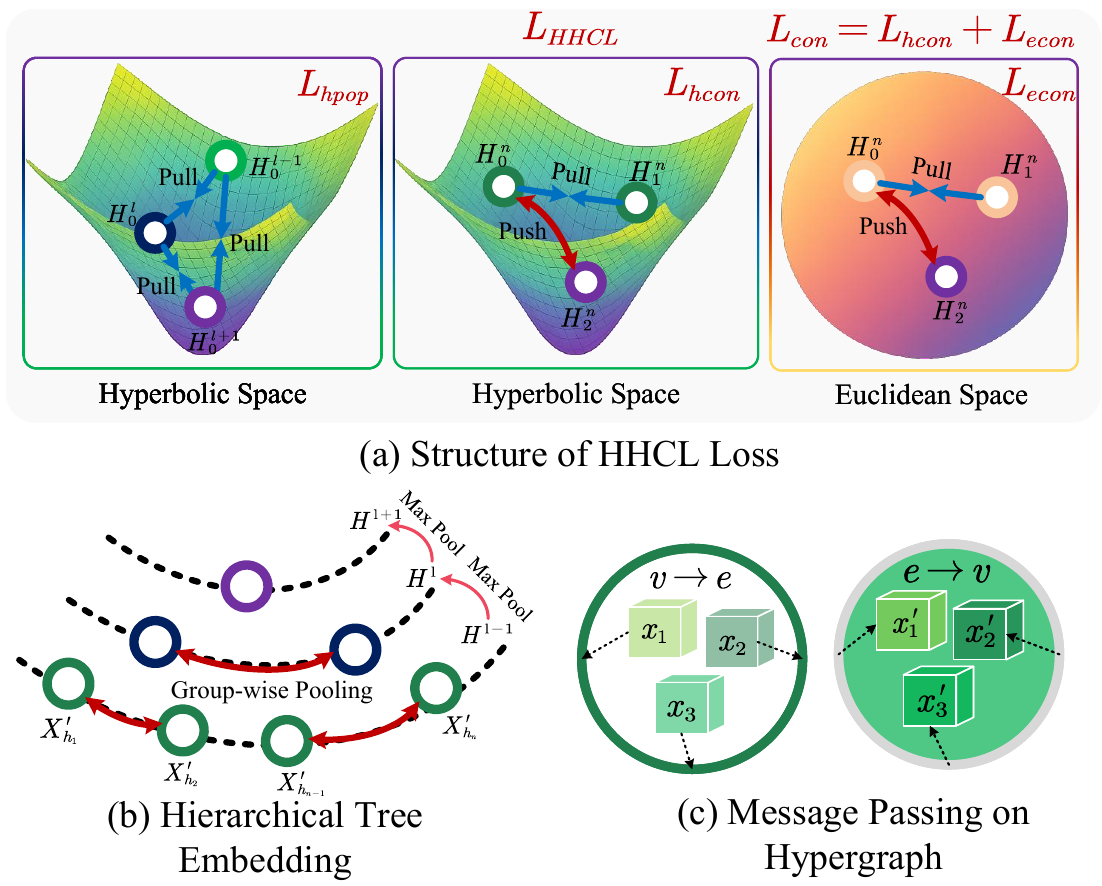} 

\caption{\textbf{Illustration of hierarchical hypergraph modeling and HHCL loss.} (a) HHCL consists of $\mathcal{L}_{hpop}$ for hierarchical consistency, $\mathcal{L}_{hcon}$ for hyperbolic contrastive learning, and $\mathcal{L}_{econ}$ for euclidean discrimination. 
(b) Hierarchical Tree Embedding constructs multi-level region representations from SAAM-produced hyperedge embeddings via group-wise max pooling.
(c) SAAM performs hypergraph message passing from tokens to regions and back.
}
\label{fig:03} 
\end{figure}

\textbf{Hypergraph Message Passing.}  
As illustrated in Fig.~\ref{fig:03} (c), given the hypergraph $\mathcal{H}$ constructed by SAAM, message passing is performed in two sequential steps to enable bidirectional information exchange between tokens and semantic regions. Let $X\in\mathbb{R}^{N\times C}$ denote the token features and $A\in\mathbb{R}^{N\times M}$ the participation matrix defining the connections between $N$ tokens and $M$ hyperedges.

\textit{(1) Node-to-hyperedge aggregation.}  
Each hyperedge feature is obtained by aggregating the token embeddings connected to it, followed by a linear transformation:
\begin{equation}
\mathbf{H}_e = (A^{\top}X)W_e\in\mathbb{R}^{M\times C},
\end{equation}
where $W_e\in\mathbb{R}^{C\times C}$ is a learnable projection matrix that refines region-level semantics.

\textit{(2) Hyperedge-to-node update.}  
The updated token features are computed by broadcasting the aggregated hyperedge representations back to their associated tokens:
\begin{equation}
X' = (A\mathbf{H}_e)W_v\in\mathbb{R}^{N\times C},
\end{equation}
where $W_v\in\mathbb{R}^{C\times C}$ projects the enhanced region information back to the token space.

To stabilize training, a learnable gate $g\in\mathbb{R}^{N}$ is applied for residual fusion:
\begin{equation}
\hat{X} = X + (g\odot X'),
\end{equation}
where $\odot$ denotes element-wise multiplication (broadcasted along the channel dimension).  
This two-step propagation (\textit{V$\rightarrow$E$\rightarrow$V}) allows each token to integrate high-order semantic context through the dynamic weighted hypergraph, yielding refined token embeddings $\hat{X}$ for subsequent classification or hierarchical contrastive learning.

\subsection{Hyperbolic Hierarchical Contrastive Loss (HHCL)}
\label{sec:3.3}

While conventional Euclidean spaces are sufficient for capturing local appearance differences, they are inherently limited in modeling global semantic structures, especially in fine-grained tasks where category hierarchies and semantic overlaps are common. To this end, we propose the HHCL, which embeds features into the lorentzian hyperbolic space and introduces dual-level supervision to preserve class-level separability and structural hierarchy.

\textbf{Hyperbolic Geometry.}
Hyperbolic spaces are naturally suited for modeling tree-like or hierarchical structures due to their exponential growth property~\cite{liu2025hyperbolic, jun2025hlformer, hyper2}. We adopt the Lorentz model, a numerically stable realization of hyperbolic geometry, which represents each point $x = [x_0, \mathbf{x}_s] \in \mathbb{R}^{d+1}$ on the upper sheet of a two-sheet hyperboloid defined by:
\begin{equation}
\mathbb{L}^{d} = \left\{ x \in \mathbb{R}^{d+1} \mid \langle x, x \rangle_{\mathcal{L}} = -1, \; x_0 > 0 \right\},
\end{equation}
where the Lorentzian inner product is $\langle x, y \rangle_{\mathcal{L}} = -x_0 y_0 + \langle \mathbf{x}_s, \mathbf{y}_s \rangle$.

To project a Euclidean feature $z \in \mathbb{R}^d$ into the hyperbolic space, we apply the exponential map at the origin:
\begin{equation}
\exp_0(z) = \left( \sqrt{1 + \|z\|^2}, z \right),
\end{equation}
which ensures the mapped point lies on the manifold. Distances in this space are computed via:
\begin{equation}
d_{\mathcal{L}}(x, y) = \text{arcosh}\left( -\langle x, y \rangle_{\mathcal{L}} \right).
\end{equation}

\textbf{Hierarchical Tree Embedding.}
As illustrated in Fig.~\ref{fig:03} (b), we construct a hierarchical tree to organize the semantic regions obtained from the dynamic hypergraph. SAAM first produces $M$ hyperedge-level region embeddings through learned token-to-hyperedge association. These hyperedge embeddings already encode semantic affinity induced by the token-to-region aggregation process. Based on them, we further construct a hierarchical tree representation for HHCL.
Formally, given the hyperedge-wise outputs
$\{\,\mathbf{X}_{h_1}',\,\mathbf{X}_{h_2}',\,\dots,\,\mathbf{X}_{h_n}'\,\}$
generated by SAAM, we denote the initial region embeddings as
$H^{1}\in\mathbb{R}^{B\times16\times d}$. The next-level representation is obtained by a deterministic hierarchical aggregation operator:
\begin{equation}
H^{\ell+1} = \mathcal{A}\big(H^{\ell}\big), \qquad \ell=1,\dots,L-1.
\end{equation}

In our implementation, $\mathcal{A}(\cdot)$ is instantiated as fixed adjacent group-wise max pooling. Specifically, for $H^{\ell}\in\mathbb{R}^{B\times N_{\ell}\times d}$, adjacent region nodes are grouped in pairs, and each parent node is computed as:
\begin{equation}
H^{\ell+1}_{b,j,r}
=
\max\left(
H^{\ell}_{b,2j,r},
H^{\ell}_{b,2j+1,r}
\right),
\end{equation}
where $b$, $j$, and $r$ denote the sample index, parent-node index, and feature dimension, respectively. The level-wise descriptors are then obtained by pooling nodes within each level, and the selected descriptors are mean-fused to form the tree representation for HHCL.

\textbf{Hybrid Contrastive Loss.}
To encourage compact intra-class clustering and inter-class separation, we perform supervised contrastive learning over the fused region representation $z_i = H_i^{\ell}$ in both Euclidean and hyperbolic spaces. Specifically, we first compute the Euclidean distance and the Lorentz hyperbolic distance between two region representations as:
\begin{equation}
D^{E}_{i,j} = \|z_i - z_j\|,
\quad
D^{H}_{i,j} = d_{\mathcal{L}}(\exp_0(z_i), \exp_0(z_j)),
\end{equation}
where $\exp_0(\cdot)$ maps the Euclidean representation to the Lorentz hyperbolic space.

Based on these two distances, we define the Euclidean contrastive loss and the hyperbolic contrastive loss as:
\begin{equation}
\mathcal{L}_{\text{econ}} =
- \sum_{i} \frac{1}{|P(i)|} \sum_{p \in P(i)}
\log \left(
\frac{\exp(-D^{E}_{i,p}/\tau)}
{\sum_{a \neq i} \exp(-D^{E}_{i,a}/\tau)}
\right),
\end{equation}
\begin{equation}
\mathcal{L}_{\text{hcon}} =
- \sum_{i} \frac{1}{|P(i)|} \sum_{p \in P(i)}
\log \left(
\frac{\exp(-D^{H}_{i,p}/\tau)}
{\sum_{a \neq i} \exp(-D^{H}_{i,a}/\tau)}
\right),
\end{equation}
where $P(i)$ denotes the set of positives sharing the same class label as $i$, and $\tau$ is a temperature hyperparameter.

The final category-wise contrastive loss is formulated as:
\begin{equation}
\mathcal{L}_{\text{con}}
=
\omega_h \mathcal{L}_{\text{hcon}}
+
\omega_e \mathcal{L}_{\text{econ}},
\end{equation}
where $\omega_h$ and $\omega_e$ control the contributions of the hyperbolic and Euclidean contrastive terms, respectively.

\textbf{Hypergraph Partial Order Preservation Loss.}
To further regularize the tree structure, we enforce that higher-level features (e.g., $H_i^{\ell+1}$) lie closer to their children (e.g., $H_i^{\ell}$) in the hyperbolic space. The partial order preservation loss (POPL) is defined as:
\begin{equation}
\mathcal{L}_{\text{hpop}} = \frac{1}{L-1} \sum_{\ell=1}^{L-1} \text{ReLU}\left( d_{\mathcal{L}}(\exp_0(H_i^{\ell+1}), \exp_0(H_i^\ell)) \right),
\end{equation}

which penalizes overly distant child-parent pairs that break semantic consistency. This encourages feature evolution to follow a smooth, hierarchical flow. The Euclidean auxiliary term is used only for the category-wise contrastive objective, where local metric discrimination helps separate different semantic categories. For the partial-order preservation objective, we keep the constraint purely in hyperbolic space, since this term is intended to preserve parent-child hierarchical relations rather than local category neighborhoods.

\textbf{Final Objective.}
As illustrated in Fig.~\ref{fig:03} (a), the complete HHCL objective consists of the hyperbolic contrastive loss, the Euclidean contrastive loss, and the hypergraph partial order preservation loss:
\begin{equation}
\mathcal{L}_{\text{HHCL}}
=
\omega_h \mathcal{L}_{\text{hcon}}
+
\omega_e \mathcal{L}_{\text{econ}}
+
\omega_p \mathcal{L}_{\text{hpop}},
\end{equation}
where $\omega_h$, $\omega_e$, and $\omega_p$ denote the weights of the three HHCL components.

\section{Experiments}

\subsection{Fine-grained Datasets.}

We incorporated four well-known public datasets for comparative analysis: CUB-200-2011~\cite{wah2011cub200}, NA-Birds~\cite{van2015nabirds}, Stanford Dogs~\cite{dogs} and Oxford Flowers-101~\cite{flower}. The CUB-200-2011 and NA-Birds datasets are dedicated to the fine-grained classification of birds, the Stanford-Dogs dataset focuses on dog species, whereas the Oxford Flowers-101 dataset focuses on flower species. These datasets present high intra-class variation and subtle inter-class differences, making them ideal benchmarks for evaluating fine-grained localization and representation learning. All experiments follow the original benchmarks' standard train/test splits.

\subsection{Implementation Details}

All experiments in this paper were conducted using PyTorch and executed on a single NVIDIA A100 graphics card. For the CUB-200-2011, the NA-Birds and the Flowers-101 dataset, we adopt the Swin-B backbone~\cite{swin} pre-trained on ImageNet-22K~\cite{21k}, while for the Stanford-Dogs dataset, we use the version pre-trained on ImageNet-1K~\cite{5206848}. All input images are resized to $448 \times 448$, processed with a sliding window of stride 14 and finally partitioned into $14 \times 14$ patches in the last stage. The embedding dimension is $\{128, 256, 512, 1024\}$, the MLP hidden dimension is $\{512, 1024, 2048, 4096\}$, and the transformer uses 12 layers with $\{4, 8, 16, 32\}$ attention heads. Our proposed SAAM module uses semantic hyperedges $M=16$. The hyperbolic terms in HHCL are computed in the Lorentz model with fixed curvature $\mathcal{K}=0.1$, and the temperature is set to $\tau=0.1$. For the total objective, the global coefficient of HHCL is set to $\alpha=1.0$. Inside HHCL, the default component weights are set to $\omega_h=0.1$, $\omega_e=0.1$, and $\omega_p=0.1$ for the hyperbolic contrastive loss, Euclidean contrastive loss, and hypergraph partial order preservation loss, respectively. The hierarchical supervision is applied across four levels with region fusion ratios of $\{16,8,4,2,1\}$.

\subsection{Comparison with the State-of-the-arts}
We conduct comprehensive comparisons with state-of-the-art fine-grained classification methods on four widely used benchmarks: CUB-200-2011~\cite{wah2011cub200}, NA-Birds~\cite{van2015nabirds}, Stanford-Dogs~\cite{dogs} and Flowers-101~\cite{flower}. The results are summarized in Tab.~\ref{tab:01}, Tab.~\ref{tab:02}, Tab.~\ref{tab:03} and Tab.~\ref{tab:r1}, respectively. Our method consistently achieves the highest accuracy across all datasets. 

As shown in Tab.~\ref{tab:01}, our model reaches 92.7\%, surpassing all existing approaches on the CUB-200-2011 dataset. In particular, it outperforms region-relation based methods such as I2-HOFI~\cite{i2o} and SR-GNN~\cite{srgnn} by +1.1\% and +0.8\%, respectively. Compared with feature-selection based methods like IELT~\cite{xu2023ielt} (91.8\%) and FAL-ViT~\cite{fal} (91.7\%), our model still yields a noticeable gain of +1.0\%, demonstrating the effectiveness of our high-order semantic aggregation strategy over token-centric alternatives.

\begin{table}[t]
    \small
    \caption{Comparison of the top-1 accuracy (\%) with the state-of-the-arts on the CUB-200-2011~\cite{wah2011cub200} dataset.}
    \centering
    \setlength{\tabcolsep}{1.6mm}{
    \begin{tabular}{l|l|l|c}
    \hline \toprule [0.7 pt]
  
    \textbf{Method} & \textbf{Publication} &\textbf{Backbone}  & \textbf{CUB-200-2011} \\ \midrule
    FDL~\cite{liu2020fdl} &AAAI 2020 & DenseNet & 89.1\\
    LOPSI~\cite{lopsi} &TMM 2021&ResNet&88.9\\
    AP-CNN~\cite{apcnn} &TIP 2021& ResNet  & 88.4\\ 
    SR-GNN~\cite{srgnn} &TIP 2022& Xception &\underline{91.9}\\
    P2P-Net~\cite{p2p} &CVPR 2022& ResNet  & 90.2\\ 
    LGTF~\cite{lgtf}&ICCV 2023&DenseNet &91.5\\
    GDSMP~\cite{gdsmp}&PR 2023&ResNet&89.9\\
    LMEPR~\cite{lmepr}&TMM 2023&RestNet &90.9\\
    I2-HOFI~\cite{i2o}&IJCV 2024&Xception &91.6 \\
    \midrule
    
    ViT-Net~\cite{vit-net}&ICML 2022&Swin-B  &91.6 \\
    TransFG~\cite{he2022transfg}&AAAI 2022&ViT-B  &91.7\\
    IELT~\cite{xu2023ielt}&TMM 2023&ViT-B  &91.8 \\
    MP-FGVC~\cite{mpFGVC}&AAAI 2024&ViT-B  &91.8  \\
    ACC-VIT~\cite{fal}&AAAI 2024&ViT-B  &91.8 \\
    FAL-ViT~\cite{fal}& TCSVT 2025&ViT-B & 91.7 \\
    TransIFC+~\cite{transifc}&TMM 2025 &Swin-B &91.0\\

    \rowcolor{gray!15}
    \textbf{$\text{H}^3$Former (Ours)} & --&Swin-B & \textbf{92.7}\\ 
    \hline \toprule [0.7 pt]
    \end{tabular}
    \label{tab:01}}
\end{table}

\begin{table}[t]
    \small
    \caption{Comparison of the top-1 accuracy (\%) with the state-of-the-arts on the NA-Birds~\cite{van2015nabirds} dataset.}
    \centering
    \setlength{\tabcolsep}{2.5mm}{
    \begin{tabular}{l|l|l|c}
    \hline \toprule [0.7 pt]
    
    \textbf{Method} &\textbf{Publication} &\textbf{Backbone}  &  \textbf{NA-Birds} \\ \midrule
    
    APIN~\cite{zhuang2020apinet}& AAAI 2020 &DenseNet &88.1\\
    CAP~\cite{cap} &AAAI 2021&Xception &91.0\\
    SR-GNN~\cite{srgnn}&TIP 2022& Xception &91.2\\
    LGTF~\cite{lgtf}& ICCV 2023&DenseNet &90.4\\
    GDSMP~\cite{gdsmp}& PR 2023&ResNet&89.0\\
    \midrule
    TransFG~\cite{he2022transfg}&AAAI 2022&ViT-B  &90.8\\
    IELT~\cite{xu2023ielt}&TMM 2023&ViT-B  &90.8 \\
    MP-FGVC~\cite{mpFGVC}&AAAI 2024&ViT-B  &91.0  \\
    ACC-VIT~\cite{accvit} &TCSVT 2025&ViT-B  &\underline{91.4}\\
    FAL-ViT~\cite{fal}&TCSVT 2025&ViT-B & 90.3 \\
    TransIFC+~\cite{transifc}& TMM 2025&Swin-B &90.9\\
    \rowcolor{gray!15}
     \textbf{$\text{H}^3$Former (Ours)} &-&Swin-B & \textbf{91.6}\\ 
    \hline \toprule [0.7 pt]
    \end{tabular}
    
    \label{tab:02}}
\end{table}

\begin{table}[t]
    \small
    \caption{Comparison of the top-1 accuracy (\%) with the state-of-the-arts on the Stanford-Dogs~\cite{dogs} dataset.}
    \centering
    \setlength{\tabcolsep}{3.3 mm}{
    \begin{tabular}{l|l|l|c}
    \hline \toprule [0.7 pt]
    
    \textbf{Method}&\textbf{Publication}  &\textbf{Backbone}  &  \textbf{Dogs}\\ \midrule
    FDL~\cite{liu2020fdl}&AAAI 2020 & DenseNet & 84.9\\
    APIN~\cite{zhuang2020apinet}&AAAI 2020& DenseNet &90.3\\
    CAR~\cite{cal}&ICCV 2021&ResNet& 88.7\\

    LGTF~\cite{lgtf}&ICCV 2023&DenseNet &92.1\\
    \midrule
    ViT-Net~\cite{vit-net}&ICML 2022&Swin-B  &\underline{93.6} \\
    TransFG~\cite{he2022transfg}&AAAI 2022&ViT-B  &92.3\\
    IELT~\cite{xu2023ielt}&TMM 2023&ViT-B  &91.8 \\
    MP-FGVC~\cite{mpFGVC}&AAAI 2024&ViT-B  &91.0  \\
    ACC-VIT~\cite{accvit}&TCSVT 2025 &ViT-B  & 92.9\\
    FAL-ViT~\cite{fal}&TCSVT 2025&ViT-B & 91.1 \\
    
    \rowcolor{gray!15}
     \textbf{$\text{H}^3$Former (Ours)}&-&Swin-B & \textbf{95.8}\\ 
    \hline \toprule [0.7 pt]
    \end{tabular}
    
    \label{tab:03}}
\end{table}

\begin{table}[t]
    \small
    \caption{Comparison of the top-1 accuracy (\%) with the state-of-the-arts on the Oxford Flowers-101~\cite{flower} dataset.}
    \centering
    \setlength{\tabcolsep}{1.7 mm}{
    \begin{tabular}{l|l|l|c}
    \hline \toprule [0.7 pt]
    
    \textbf{Method} &\textbf{Publication} & \textbf{Backbone}  &  \textbf{Flowers101} \\ \midrule

    PBC~\cite{pbc}&TMM 2016&GoogleNet&96.1\\
    InAct~\cite{interact}&CVPR 2016&VGG&96.4\\
    SJFT~\cite{sjft}&CVPR 2017&ResNet&97.0\\
    OPAM~\cite{opam}&TIP 2017&VGG&97.1\\
    DSTL~\cite{dstl}&CVPR 2018&Inceaption-v3&97.6\\
    MGE~\cite{mge} &CVPR 2019& ResNet&95.9\\
    Cos.Ls~\cite{cos}&WACV 2020&ResNet-50&97.2\\
    PMA~\cite{pma}&TIP 2020&VGG& 97.4\\
    MCL~\cite{mcl}&TIP 2020&Bilinear CNN&97.7\\
    CAP~\cite{cap}&AAAI 2021&Xception&97.7\\
    SR-GNN~\cite{srgnn}&TIP 2022&Xception&97.9\\
    I2-HOFI~\cite{i2o}&IJCV 2024&Xception &\underline{99.0} \\

    \rowcolor{gray!15}
     \textbf{$\text{H}^3$Former (Ours)}&- &Swin-B & \textbf{99.7}\\ 
    \hline \toprule [0.7 pt]
    \end{tabular}
    
    \label{tab:r1}}
\end{table}
As shown in Tab.~\ref{tab:02}, our method attains 91.6\%, achieving a +0.6\% improvement over the multimodal prompting method MP-FGVC~\cite{mpFGVC} (91.0\%) on the NA-Birds dataset. Notably, despite using the same Swin-B backbone, our framework outperforms TransIFC+~\cite{transifc} by +0.7\%, highlighting the effectiveness of $\text{H}^3$Former. 

As shown in Tab.~\ref{tab:03}, our model reaches 95.8\% on the Stanford-Dogs dataset. This surpasses previous leading methods such as FAL-ViT~\cite{fal} (91.1\%) and ACC-ViT~\cite{accvit} (92.9\%) by large margins of +4.7\% and +2.9\%, respectively. The substantial improvement in this challenging dataset with high intra-class variance further verifies our proposed $\text{H}^3$Former's robustness and generalization capability in FGVC domains. This improvement stems from the SAAM, which identifies class-relevant tokens and captures their high-order semantic correlations via hypergraph modeling, resulting in more structured and discriminative representations.

To further demonstrate the generalization ability of our proposed H\textsuperscript{3}Former, we evaluate it on the widely used Oxford Flowers-101 dataset. As shown in Tab.~\ref{tab:r1}, our method achieves a top-1 classification accuracy of \textbf{99.7\%}, setting a new state-of-the-art on this benchmark. Compared to classical convolutional backbones such as GoogleNet, VGG, and ResNet used in earlier works like PBC~\cite{pbc}, OPAM~\cite{opam}, and MGE~\cite{mge}, our model improves accuracy by over +3.0\%, showing that H\textsuperscript{3}Former benefits from both the hierarchical representation of Swin Transformer and the semantic structuring capability of hypergraphs. More importantly, our model outperforms recent strong fine-grained baselines.~\ie, compared to SR-GNN~\cite{srgnn} and I2-HOFI~\cite{i2o}, which leverage graph-based relational modeling and achieve 97.9\% and 99.0\% accuracy, our model yields relative gains of +1.8\% and +0.7\%, respectively. This highlights that our semantic-aware token-to-region aggregation strategy better captures category-specific structures in dense visual scenes like flowers, where visual differences are subtle and often localized. In addition, feature channel enhancement-based methods such as CAP~\cite{cap} and MCL~\cite{mcl} achieve 97.7\%, whereas H\textsuperscript{3}Former surpasses them by a large margin of +2.0\%. This improvement can be attributed to two key factors: (1) our SAAM module adaptively groups fine-grained semantic tokens into high-order regions, and (2) our HHCL enhances intra-class consistency and inter-class separation in the embedding space.

Overall, these results further validate the robustness and scalability of our approach across diverse domains. The consistent performance demonstrates that H\textsuperscript{3}Former is effective for animals or plants and excels in complex multi-instance scenes with high intra-class variation and low inter-class separability.

\begin{figure*}[t]
\centering 

\includegraphics[width=1\linewidth]{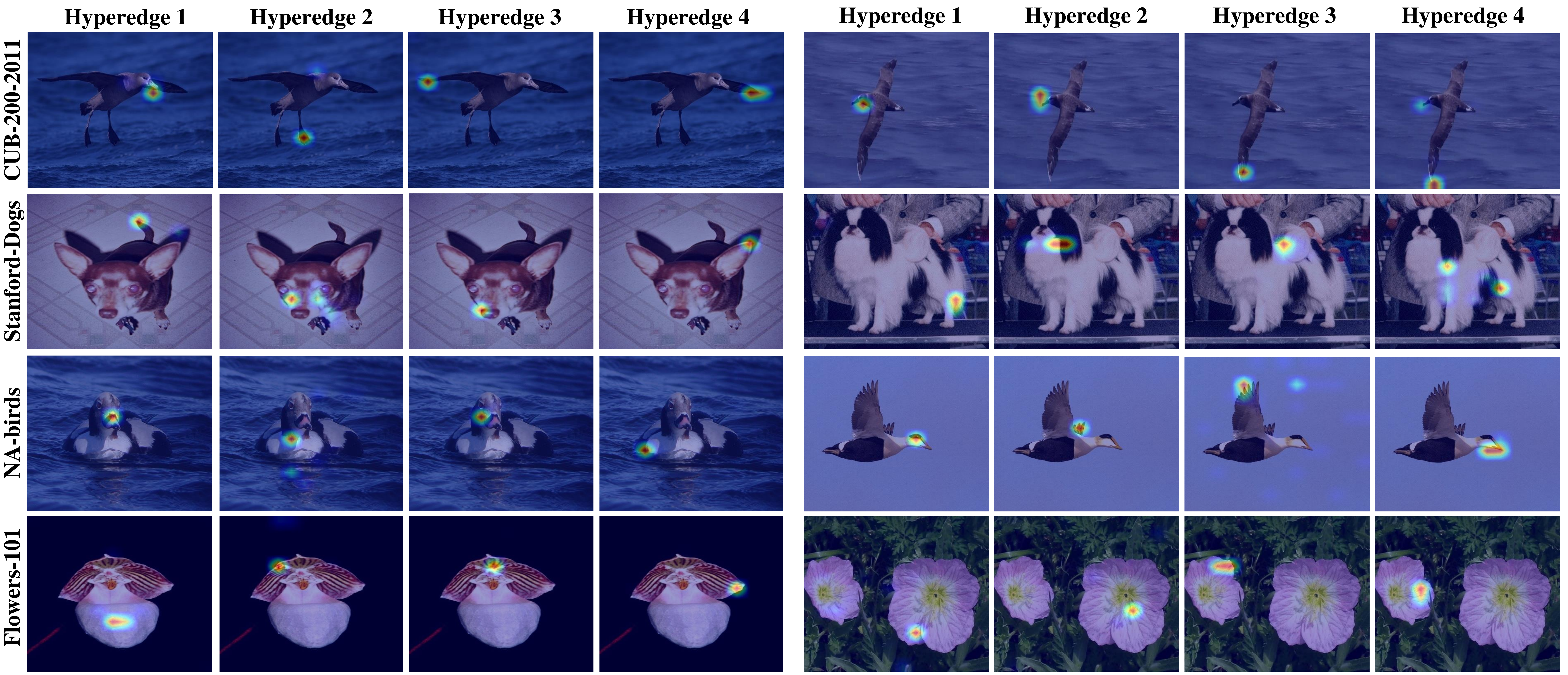} 

\caption{\textbf{Visualization of hyperedges in $\text{H}^3$Former.} Each row shows two input images from same dataset and the activation maps of four hyperedges. Each hyperedge captures a distinct semantic region,~\eg, the beak, wings, or feet of the bird. This illustrates the semantic-aware and complementary nature of our $\text{H}^3$Former.}

\label{fig:sp1} 
\end{figure*}

\subsection{Visualization}

To further illustrate the interpretability and semantic structure modeling capability of the proposed H\textsuperscript{3}Former, we present qualitative visualizations of learned hyperedges in Fig.~\ref{fig:sp1}. Each row corresponds to two sample images from the four datasets, and each column visualizes the token-level activation associated with one of the learned hyperedges. We visualize the learned hypergraph by mapping token-to-hyperedge weights into spatial heatmaps overlaid on the input images. Each hyperedge highlights distinct semantic regions, revealing how the model adaptively groups correlated tokens into meaningful structures.

From the visualizations, we observe that different hyperedges often respond to object-related or part-related discriminative regions, such as regions around the head, beak, wings, feet, or tail. These observations provide qualitative evidence that the hypergraph-based semantic aggregation module can group correlated tokens into meaningful region-level semantic structures without using explicit part annotations or predefined proposals.

\begin{figure}[t]
\centering 
\includegraphics[width=1\linewidth]{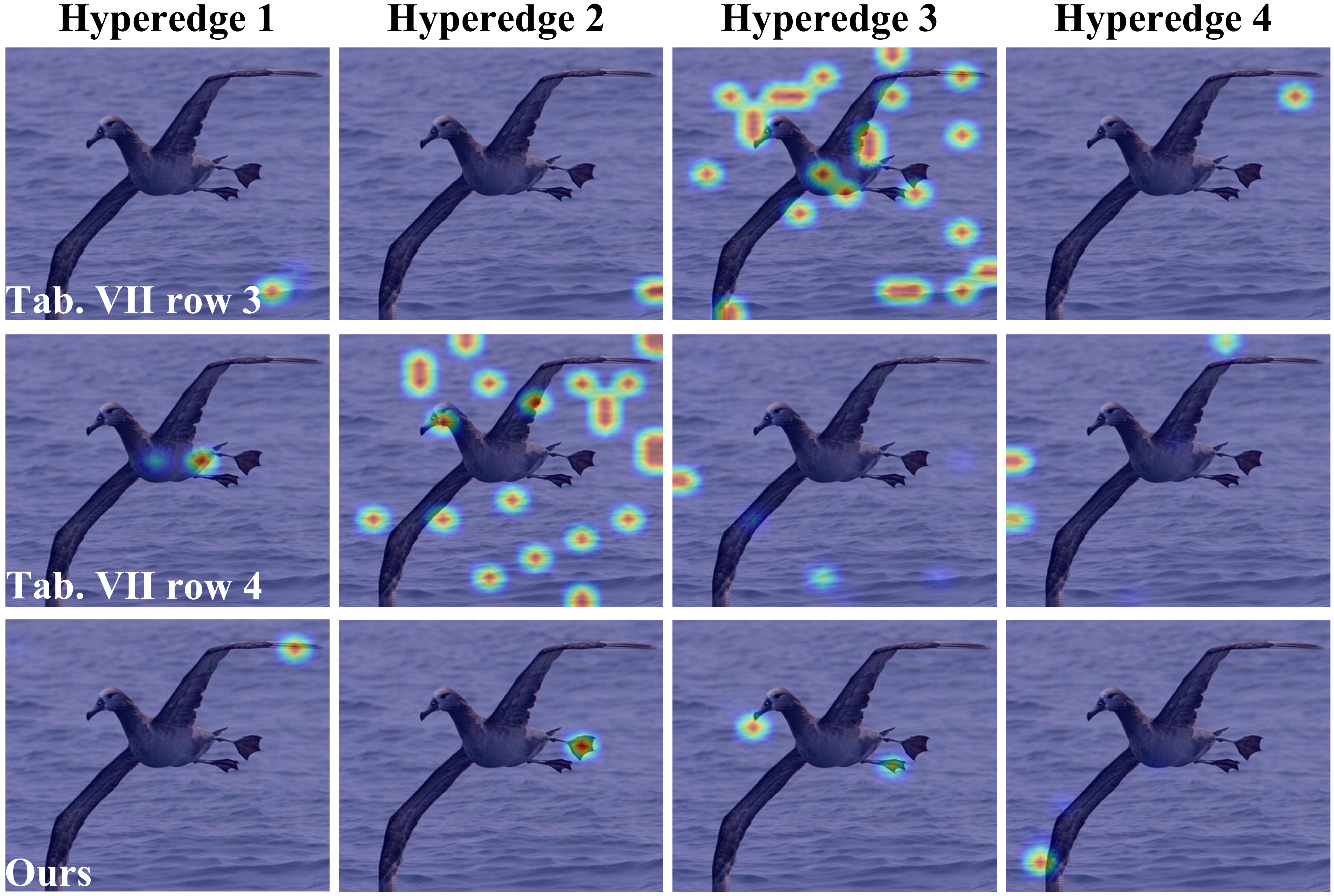} 

\caption{\textbf{Visualization of hyperedges learned with different loss designs.} 
Each column corresponds to one hyperedge, and each row shows the token-level activation maps of the models from Tab.~\ref{tab:05} (rows 3 and 4) and our proposed method. 
Replacing our HHCL with alternative hyperbolic loss functions leads to less compact or inconsistent semantic grouping, 
while our HHCL tightly aligns with the hypergraph structure, producing clearer and more coherent semantic regions.
}

\vspace{-0.2cm}
\label{fig:sp2} 
\end{figure}

Interestingly, the activations of the hyperedges are not redundant but complementary. While some hyperedges capture dominant features like the bird's head or torso, others attend to more subtle or context-specific details such as leg orientation or feather curvature. This complementary behavior enhances the model's ability to capture high-order semantic cues across different spatial locations, which is especially crucial for recognizing fine-grained differences. Moreover, the consistency across different images (even under pose variation or occlusion) highlights the robustness of our semantic-aware aggregation. Unlike traditional attention mechanisms that may focus inconsistently across samples, the hypergraph formulation allows H\textsuperscript{3}Former to form stable, interpretable region groupings aligned with object structure. These visual results support our key intuition that adaptive hyperedges can serve as soft region-level semantic grouping units, enabling structured representation learning for fine-grained categorization.

As shown in Fig.~\ref{fig:sp2}, we visualize the token-to-hyperedge activations of models trained with different loss variants to further analyze the relationship between the hypergraph structure and the proposed hyperbolic contrastive design. Each hyperedge focuses on distinct semantic regions of the object, such as the head, wings, or feet. Compared with alternative hyperbolic losses, our HHCL produces more compact, part-consistent activations, whereas the others tend to yield scattered or overlapping responses that blur regional boundaries. This indicates that HHCL establishes a stronger alignment between the hypergraph structure and semantic aggregation. Compared with alternative hyperbolic losses, our HHCL tends to produce more compact and less overlapping region responses, while the other variants often yield scattered or ambiguous activations.

\begin{table}[t!]
\centering
\small
\caption{\textbf{Post-hoc set-level alignment with CUB part groups.}
CUB part landmarks are used only for evaluation, not for training. We group the 15 landmarks into five coarse part groups and evaluate whether the learned hyperedge set contains part-related responses on the original token grid.}
\label{tab:cub_part_group_alignment}
\setlength{\tabcolsep}{5.0pt}
\begin{tabular}{lcc}
\Xhline{1.2pt}
\textbf{Method} & \textbf{Best-1 PAS} $\uparrow$ & \textbf{Best-2 PAS} $\uparrow$ \\
\hline
Shuffled Response & 0.735  & 0.708  \\
H$^3$Former Hyperedges & \textbf{0.930} & \textbf{0.909} \\
\Xhline{1.2pt}
\end{tabular}
\end{table}
\textbf{Post-hoc Set-level Alignment with CUB Part Groups.} Although H$^3$Former is not trained with part annotations, we further examine whether the learned hyperedge set is related to CUB part landmarks. Since the hyperedges are image-adaptive semantic regions rather than explicit part detectors, we do not enforce a one-to-one correspondence between hyperedge IDs and predefined anatomical parts. Instead, we conduct a set-level alignment analysis.

Specifically, we group the CUB landmarks into five coarse part groups: head, torso, wing, leg, and tail. For each image, the visible part landmarks are mapped to the original token grid, and a small neighborhood is constructed around each landmark to account for the spatial granularity of token-level responses. For each hyperedge response map, we compute a percentile response map on the token grid. Given a part-group neighborhood, we average the percentile responses inside this region to measure how strongly the hyperedge responds to this part-related area.

Because fine-grained recognition often relies on one or two discriminative regions rather than all object parts, we report Best-1 and Best-2 Part Alignment Scores (PAS). Best-1 PAS selects the highest part-group response among all hyperedges and visible part groups for each image, while Best-2 PAS averages the top two responses from different part groups. We compare the learned hyperedge responses with a spatially shuffled baseline, which preserves the response value distribution but destroys the spatial correspondence with object parts. As shown in Tab.~\ref{tab:cub_part_group_alignment}, H$^3$Former achieves higher Best-1 and Best-2 PAS than the shuffled response baseline. This indicates that the learned hyperedge set tends to contain part-related discriminative responses without using part-level supervision.

\begin{figure}[t!]
\centering 

\includegraphics[width=1\linewidth]{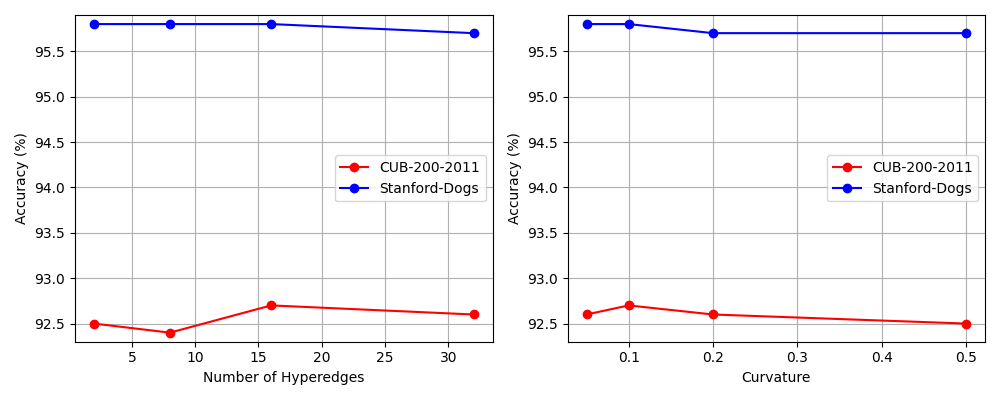} 

\caption{\textbf{Influence of hyperparameters on classification accuracy on the CUB-200-2011 and Stanford-Dogs datasets.} (a) Accuracy curves with the numbers of hyperedges $M$. (b) Accuracy curves with he curvature $\mathcal{K}$ in the Lorentzian embedding.}
\label{fig:04} 
\end{figure}
\begin{figure}[t!]
\centering 
\includegraphics[width=1\linewidth]{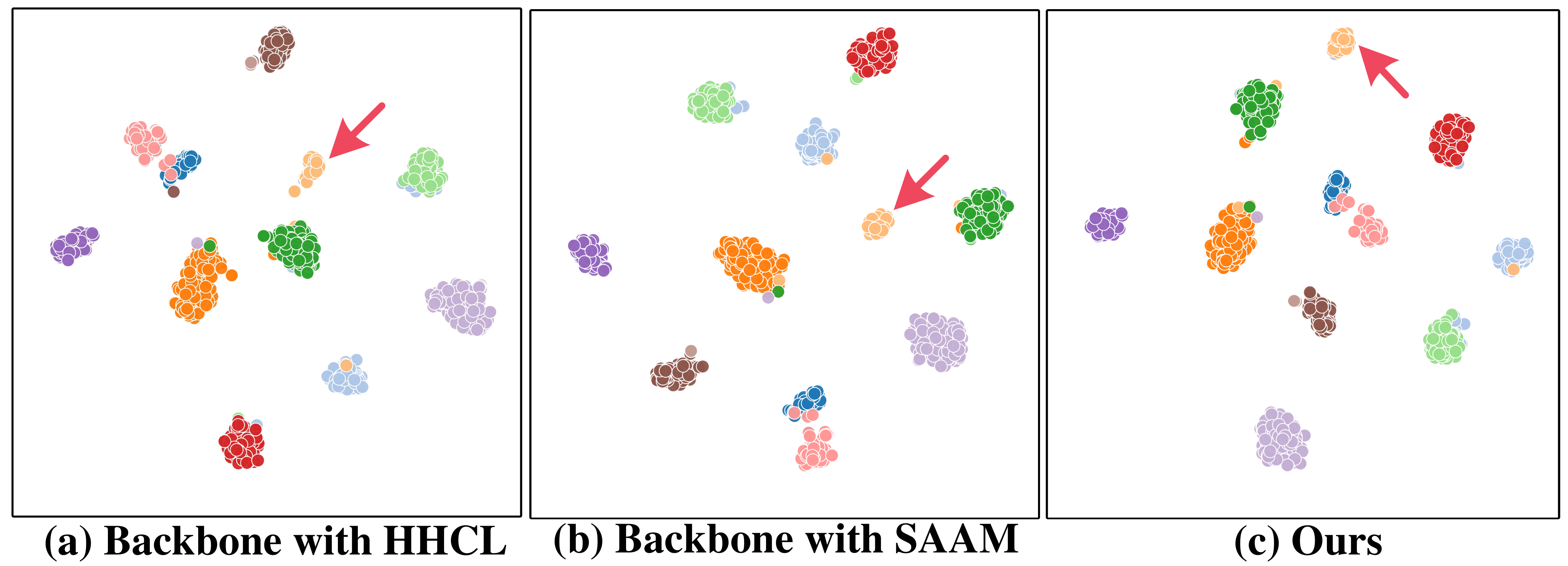}

\caption{\textbf{t-SNE visualizations on the Stanford-Dogs dataset.} (a) Features with HHCL only. (b) Features with SAAM only. (c) Features incorporating both SAAM and HHCL, demonstrates clearer clustering and enhanced inter-class separability.}
\vspace{-0.1cm}
\label{fig:06} 
\end{figure}

\begin{table}[t]
\begin{center}
\small{
\caption{Ablation studies on CUB-200-2011~\cite{wah2011cub200} and Stanford-Dogs~\cite{dogs} dataset. \cmark denotes the component is added. \xmark denotes the component is removed.}
\setlength{\tabcolsep}{3.6mm}{
  \begin{tabular}{ ccl l }
  
\Xhline{1.2pt}
    \centering

    \textbf{SAAM}& \textbf{HHCL}& \textbf{CUB-200-2011}& \textbf{Stanford-Dogs} \\
    \hline
    \xmark & \xmark & $90.9^{\downarrow 1.8}$ & $91.1^{\downarrow 4.7}$\\
    \xmark & \cmark & $91.2^{\downarrow 1.5}$ & $92.6^{\downarrow 3.2}$ \\
    \cmark  & \xmark & $92.5^{\downarrow 0.2}$ & $95.2^{\downarrow 0.6}$\\
    \rowcolor{gray!15}
    \cmark  & \cmark  &\textbf{92.7} & \textbf{95.8}\\
    \hline

    \Xhline{1.2pt}
  \end{tabular}
  }

\label{tab:04}}
\end{center} 
\end{table}

\begin{table}[t!]
\small
\caption{\textbf{Ablation on different HHCL component weights.}
The coefficient of $\mathcal{L}_{\text{CE}}$ is fixed to 1.0 for all settings. We vary the weights of the hyperbolic contrastive loss $\mathcal{L}_{hcon}$, Euclidean contrastive loss $\mathcal{L}_{econ}$, and hypergraph partial order preservation loss $\mathcal{L}_{hpop}$ on the Stanford-Dogs dataset.}
\centering
\setlength{\tabcolsep}{2.2mm}{
\begin{tabular}{cccl}
\hline \toprule [0.7 pt]
$\omega_h$ for $\mathcal{L}_{hcon}$ 
& $\omega_e$ for $\mathcal{L}_{econ}$ 
& $\omega_p$ for $\mathcal{L}_{hpop}$ 
& Stanford-Dogs \\
\midrule
0.0  & 0.0  & 0.0  & $95.2^{\downarrow 0.6}$ \\
0.1  & 0.0  & 0.0  & $95.5^{\downarrow 0.3}$ \\
0.05 & 0.0  & 0.0  & $95.3^{\downarrow 0.5}$ \\
0.2  & 0.0  & 0.0  & $95.4^{\downarrow 0.4}$ \\
0.1  & 0.05 & 0.0  & $95.5^{\downarrow 0.3}$ \\
0.1  & 0.1  & 0.0  & $95.6^{\downarrow 0.2}$ \\
0.1  & 0.2  & 0.0  & $95.1^{\downarrow 0.7}$ \\
0.1  & 0.0  & 1.0  & $95.6^{\downarrow 0.2}$ \\
\rowcolor{gray!15}
0.1  & 0.1  & 0.1  & \textbf{95.8} \\
0.5  & 0.5  & 0.1  & $95.6^{\downarrow 0.2}$ \\
0.1  & 0.1  & 0.05 & $95.5^{\downarrow 0.3}$ \\
0.1  & 0.1  & 0.5  & $95.2^{\downarrow 0.6}$ \\
0.1  & 0.1  & 0.2  & $95.4^{\downarrow 0.4}$ \\
\hline \toprule [0.7 pt]
\end{tabular}
\label{tab:r2}
}
\vspace{-0.3cm}
\end{table}
\begin{table}[t]
\begin{center}
\small{
\caption{\textbf{Ablation on the Context Generation Module.} 
We evaluate the contribution of max pooling, average pooling, and attention in CGM on the CUB-200-2011 and Stanford-Dogs datasets. All other settings are kept unchanged.}
\setlength{\tabcolsep}{2.6mm}{
  \begin{tabular}{ccc}
\Xhline{1.2pt}
    \textbf{CGM Variant} & \textbf{CUB-200-2011} & \textbf{Stanford-Dogs} \\
    \hline
    w/o Max Pool & 92.5 & 95.5 \\
    w/o Avg Pool & 92.6 & 95.7 \\
    w/o Attention & 92.5 & 95.6 \\
    \rowcolor{gray!15}
    Ours & \textbf{92.7} & \textbf{95.8} \\
\Xhline{1.2pt}
  \end{tabular}
}
\label{tab:ablation_cgm}
}
\end{center}
\end{table}
\begin{table}[t]
\begin{center}
\small{
\caption{\textbf{Backbone generalization analysis.}
We evaluate the effectiveness of H$^3$Former with Swin-B and ViT-B/16 backbones on CUB-200-2011 and Stanford-Dogs.}
\setlength{\tabcolsep}{2.6mm}{
  \begin{tabular}{llll}
  
\Xhline{1.2pt}
    \centering
    \textbf{Backbone} & \textbf{Method} & \textbf{CUB-200-2011} & \textbf{Stanford-Dogs} \\
    \hline

    \multirow{2}{*}{Swin-B}
    & Baseline & 90.9 & 91.1 \\
    & Ours     & $\textbf{92.7}^{\uparrow 1.8}$ & $\textbf{95.8}^{\uparrow 4.7}$ \\
    \midrule
    \multirow{2}{*}{ViT-B/16}
    & Baseline & 90.3 & 90.6 \\
    & Ours     & $\textbf{91.1}^{\uparrow 0.8}$ & $\textbf{91.6}^{\uparrow 1.0}$ \\

    \hline
\Xhline{1.2pt}
  \end{tabular}
  }
\label{tab:backbone_scalability}}
\end{center}
\end{table}

\begin{table}[t]
\begin{center}
\small{
\caption{\textbf{Quantitative analysis of feature-space organization.}
We compute local neighborhood consistency and class-center retrieval accuracy in the original feature space before t-SNE projection.}
\setlength{\tabcolsep}{2.2mm}{
  \begin{tabular}{llccc}
\Xhline{1.2pt}
    \textbf{Dataset} & \textbf{Method} & \textbf{NN@1 $\uparrow$} & \textbf{NN@5 $\uparrow$} & \textbf{Center@1 $\uparrow$} \\
    \hline

    \multirow{2}{*}{CUB}
    & Backbone    & 0.8832 & 0.8571 & 0.9218 \\
    & H$^3$Former & \textbf{0.8876} & \textbf{0.8590} & \textbf{0.9304} \\
    \midrule

    \multirow{2}{*}{Dogs}
    & Backbone    & 0.9119 & 0.9034 & 0.9309 \\
    & H$^3$Former & \textbf{0.9374} & \textbf{0.9343} & \textbf{0.9542} \\

\Xhline{1.2pt}
  \end{tabular}
}
\label{r11}}
\end{center}
\end{table}

\begin{table}[t]
\begin{center}
\small{
\caption{\textbf{Ablation on the hypergraph mapping dimension.} 
We evaluate the effect of different hypergraph mapping dimensions in the proposed SAAM on the CUB-200-2011 and Stanford-Dogs datasets. All other settings are kept unchanged.}
\setlength{\tabcolsep}{2.6mm}{
  \begin{tabular}{ccc}
\Xhline{1.2pt}
    \textbf{Mapping Dim.} & \textbf{CUB-200-2011} & \textbf{Stanford-Dogs} \\
    \hline
    256 & 92.3 & 95.5 \\
    512 & 92.5 & 95.7 \\
    \rowcolor{gray!15}
    768 & \textbf{92.7} & \textbf{95.8} \\
    1024 & 92.7 & 95.7 \\
\Xhline{1.2pt}
  \end{tabular}
}
\label{tab:ablation_mapping_dim}}
\end{center}
\end{table}

\subsection{Ablation Studies}

 \textbf{Components Ablation.}
As shown in Tab.~\ref{tab:04}, removing both modules leads to significantly lower performance (90.9\% on CUB-200-2011 and 91.1\% on Stanford-Dogs). Introducing HHCL alone improves performance to 91.2\% and 92.6\%, while incorporating only SAAM yields a more substantial gain (92.5\% and 95.2\%). When both modules are enabled, our model achieves the best results, 92.7\% on the CUB-200-2011 dataset and 95.8\% on the Stanford-Dogs dataset, demonstrating that SAAM and HHCL are complementary in enhancing FGVC.

To further understand the contribution and interplay of different loss components in our proposed HHCL, we conduct a series of ablation experiments by varying the weighting ratios of each sub-loss. As summarized in Tab.~\ref{tab:r2}, the total objective consists of the standard cross-entropy loss $\mathcal{L}_{\text{CE}}$, and three components in $\mathcal{L}_{\text{HHCL}}$: the hyperbolic contrastive loss $\mathcal{L}_{hcon}$, the Euclidean contrastive loss $\mathcal{L}_{econ}$, and the hypergraph partial order preservation loss $\mathcal{L}_{hpop}$. We observe that using only $\mathcal{L}_{\text{CE}}$ results in relatively lower accuracy (95.2\%), while the inclusion of any individual HHCL component provides noticeable performance gains. For instance, adding only $\mathcal{L}_{hcon}$ boosts performance to 95.5\%, and jointly using all three with equal weights (~\ie, 0.1 for each) further improves accuracy to \textbf{95.8\%}, which is the best result among all tested settings.

Interestingly, increasing the weights of any single sub-loss beyond 0.1 (e.g., 0.5) does not lead to further gains and may even slightly reduce accuracy (95.6\%), indicating a potential imbalance in optimization when overemphasizing a single geometric constraint. This suggests that while each component is beneficial, their contributions are most effective when balanced, reflecting their complementary roles in structuring the feature space. In particular, $\mathcal{L}_{hcon}$ simultaneously promotes inter-class discrimination and intra-class cohesion in Euclidean space by pulling together positive pairs and pushing away negatives. $\mathcal{L}_{econ}$ is a regularizer in hyperbolic space to suppress semantic drift and enforce locally compact class-wise distributions through entropy minimization. $\mathcal{L}_{hpop}$ promotes a tree-like hierarchy to preserve semantic orders.

These results validate our design of HHCL as a multi-faceted regularizer that enhances discriminative representation learning in hyperbolic space when applied in an appropriately weighted manner.

Table~\ref{tab:ablation_cgm} evaluates the contribution of different components in CGM. We observe that removing max pooling, average pooling, or attention consistently degrades the performance on both CUB-200-2011 and Stanford-Dogs. This demonstrates that these components provide complementary contextual cues: max pooling emphasizes salient local responses, average pooling captures global contextual statistics, and attention adaptively reweights discriminative token relations. The complete CGM achieves the best performance, confirming its effectiveness in generating informative context for hypergraph-based semantic aggregation.

Table~\ref{tab:backbone_scalability} further evaluates the scalability of H$^3$Former under different backbones. In addition to Swin-B, we instantiate the proposed method on a standard ViT-B/16 backbone to provide a cleaner comparison with ViT-based FGVC methods. Since standard ViT does not naturally provide multi-stage hierarchical features as Swin does, we apply H$^3$Former to the token representations before the final classification layer, following the common ViT-based FGVC setting. As shown in Tab.~\ref{tab:backbone_scalability}, our method improves the ViT-B/16 baseline by 0.8\% on CUB-200-2011 and 0.9\% on Stanford-Dogs, indicating that the proposed token-to-region semantic modeling strategy is not limited to Swin-B and can generalize to standard ViT backbones.

To quantitatively evaluate the representation quality beyond t-SNE visualization, 
we measure local neighborhood consistency (NN@K) and class-center retrieval accuracy (Center@1) in the original feature space before projection. These metrics better reflect fine-grained representation quality by evaluating whether samples are organized around semantically consistent local neighborhoods and correct class prototypes. As shown in Tab.~\ref{r11}, H$^3$Former consistently outperforms the backbone on all metrics across both datasets, demonstrating that the proposed method produces more discriminative and semantically structured feature representations for fine-grained recognition.

\begin{table}[t!]
\begin{center}
\small{
\caption{Ablation studies of different aggregation strategies and hyperbolic loss in $\text{H}^3$Former. Our approach effectively captures semantic associations between tokens. }
\setlength{\tabcolsep}{1.mm}{
  \begin{tabular}{ lll }
  
\Xhline{1.2pt}
    \centering

    \textbf{Methods}& \textbf{CUB-200-2011}& \textbf{Stanford-Dogs} \\
    \hline
    HHCL w. HGNN~\cite{softhgnn} &$92.3^{\downarrow 0.4}$ &$95.4^{\downarrow 0.4}$ \\
    HHCL w. GNN~\cite{srgnn} &$92.2^{\downarrow 0.5}$ &$95.2^{\downarrow 0.6}$\\
    \midrule
    SAAM w. Hyperbolic Loss~\cite{wacv} &$92.0^{\downarrow 0.7}$ &$95.1^{\downarrow 0.7}$\\
    SAAM w. Hyperbolic Loss~\cite{hyper2}  &$92.3^{\downarrow 0.5}$ &$95.0^{\downarrow 0.8}$\\
    \rowcolor{gray!15}
    Ours (SAAM + HHCL) &\textbf{92.7} & \textbf{95.8}\\
    \hline

    \Xhline{1.2pt}
  \end{tabular}
  }
\label{tab:05}}
\end{center} 
\end{table}
\begin{table}[t!]
\centering
\caption{The computational cost analysis of our method with recent transformer-based works.
The input size denotes the height and width of the input image.}
\label{tab:06}
\small
\setlength{\tabcolsep}{1.0mm}{
\begin{tabular}{l l c c c c}
\Xhline{1.2pt}
\textbf{Method} & \textbf{Backbone} & \textbf{\makecell[c]{Input \\Size}} & \textbf{\makecell[c]{Param. \\(M)}} & \textbf{\makecell[c]{FLOPs \\(G)}} & \textbf{\makecell[c]{Memory \\(GB)}} \\
\midrule
ViT~\cite{vaswani2017attention}                 & ViT-B/16 & 448 & 86.4 &  78.5 & 1.5 \\
RAMS-Trans~\cite{hu2021rams}         & ViT-B/16 & 448 & 86.4 & 157.4 & 2.5 \\
TransFG~\cite{he2022transfg}         & ViT-B/16 & 448 & 86.4 & 130.2 & 1.4 \\
IELT~\cite{xu2023ielt}               & ViT-B/16 & 448 & 93.5 &  73.2 & 1.2 \\
ACC-ViT~\cite{accvit}          & ViT-B/16 & 448 & 87.0 & 162.9 & 2.0 \\
\midrule
Swin-Base~\cite{swin}          & Swin-B   & 384 & 87.1 &  47.2 & 1.2 \\
ViT-Net~\cite{vit-net}          & Swin-B   & 448 & 92.2 &  65.6 & 1.4 \\
Ours                          & Swin-B   & 448 & 96.6 & 61.2  & 1.7 \\
\hline
\Xhline{1.2pt}
\end{tabular}}
\end{table}

\textbf{Hyperparameters Ablation.}

We further investigate the impact of two critical hyperparameters: the number of hyperedges $M$ in high-order token aggregation and the curvature $\mathcal{K}$ in the Lorentzian embedding space. As shown in Fig.~\ref{fig:04} (left), increasing $M$ from 2 to 16 consistently improves performance on both CUB-200-2011 and Stanford-Dogs, with the best results observed at $M{=}16$. This highlights the importance of a balanced number of semantic groupings—insufficient hyperedges may fail to capture high-order relationships, whereas excessive ones could introduce redundancy or noise (increasing $M$ to 32). On the right of Fig.~\ref{fig:04}, we examine the effect of curvature $\mathcal{K}$ by varying it from 0.05 to 0.5. Both datasets reach peak accuracy at $\mathcal{K}{=}0.1$, suggesting that a moderate negative curvature better captures global semantic structure while preserving optimization stability. Further increasing $\mathcal{K}$ slightly degrades performance, likely due to excessive geometric distortion in hyperbolic space.

Table~\ref{tab:ablation_mapping_dim} reports the effect of different hypergraph mapping dimensions in SAAM. As the mapping dimension increases from 256 to 768, the performance consistently improves on both CUB-200-2011 and Stanford-Dogs, indicating that a larger embedding space is beneficial for modeling high-order relations among selected tokens. When the dimension is further increased to 1024, the performance on CUB-200-2011 remains unchanged while that on Stanford-Dogs slightly drops. This suggests that an excessively large mapping dimension does not bring additional benefits and may introduce redundancy. Therefore, we adopt a mapping dimension of 768 in all experiments as a good trade-off between effectiveness and efficiency.

 \textbf{Feature Separability Ablation.}
To gain insight into how SAAM and HHCL influence the feature space, we visualize the learned embeddings using t-SNE under three configurations: (a) backbone with HHCL only, (b) backbone with SAAM only, and (c) our whole model. As shown in Fig.~\ref{fig:06}, adding HHCL (a) and SAAM (b) introduces more precise class boundariesnd promotes more compact clustering. When both modules are used (c), the resulting feature space exhibits the most distinct and well-separated clusters, validating the synergy between geometric supervision and semantic-aware aggregation.

 \textbf{Aggregation Strategies and Hyperbolic Loss Ablation.}
Tab.~\ref{tab:05} reports ablation results by replacing either the proposed SAAM 
or the HHCL loss with alternative designs. 
In the first group, we substitute SAAM with two representative graph-based modules: 
HGNN~\cite{softhgnn} from SoftHGNN and a standard GNN block~\cite{srgnn}. 
Both variants achieve reasonable performance but fall behind our design, 
indicating that conventional pairwise or fixed hypergraph aggregation 
is less effective in capturing fine-grained semantic associations than 
our adaptive soft hypergraph construction. In the second group, we keep SAAM but replace HHCL with two existing hyperbolic learning objectives~\cite{wacv, hyper2}. 
Although these alternatives improve discriminability compared with using cross-entropy alone, they lack explicit hierarchical modeling and yield inferior results compared to HHCL. 

Overall, our full model (SAAM + HHCL) consistently outperforms all variants, 
demonstrating the effectiveness of combining adaptive semantic aggregation 
with hierarchical hyperbolic contrastive supervision.

 \textbf{Computation Cost Analysis.}
We further compare the computational complexity of our method with recent transformer-based approaches, as summarized in Tab.~\ref{tab:06}. All methods are evaluated under the same input resolution to ensure a fair comparison. For ViT-based architectures, models such as RAMS-Trans~\cite{hu2021rams}, and ACC-ViT~\cite{accvit} exhibit high FLOPs and memory consumption due to global token interactions. Compared to current Swin-based architecture methods, \eg, ViT-Net~\cite{vit-net}, our semantic region aggregation design effectively enhances feature representation without incurring excessive computational cost.

\section{Conclusion}

In this paper, we proposed $\text{H}^3$Former, a novel framework addressing critical challenges in FGVC. The proposed SAAM dynamically constructs a weighted hypergraph to progressively aggregate visual tokens into structured and semantically coherent regions. Building upon these representations, the HHCL further enhances discriminability by enforcing hierarchical contrastive constraints within two spaces. By integrating semantic-aware region construction with geometry-aware representation learning, $\text{H}^3$Former successfully captures region-level semantic structures, effectively bridging the gap between local appearance cues and holistic object understanding. Extensive experiments on multiple FGVC benchmarks demonstrate the superior performance and generalization capabilities of our approach compared to state-of-the-art methods. 

\section{Limitations and Future Work}

The current design of H$^3$Former is centered on weakly supervised token-to-region representation learning for FGVC. The learned hyperedges act as soft region-level semantic grouping units, rather than explicit part detectors. Thus, while they can capture part-related discriminative regions, they are not intended to serve as supervised part localization outputs. Future work may incorporate part-level annotations, semantic constraints, or weakly supervised localization signals to make hyperedge formation more controllable and interpretable. In addition, SAAM is naturally suited to multi-stage feature representations. Hierarchical backbones such as Swin provide multi-scale features that align well with the context aggregation design, whereas plain ViT backbones require a less direct adaptation. Designing a more backbone-agnostic multi-level token construction mechanism is therefore a promising future direction.

Beyond FGVC, the proposed token-to-region aggregation framework also has potential for broader structured visual understanding tasks. Its ability to organize correlated local cues into high-order semantic regions could benefit action recognition~\cite{higcin,progressive,mvp,spatio,tang2023m3net}, cross-modal event localization~\cite{end,bridging}, and token compression~\cite{vision,feng2026see,shen2026efficient} in vision-language models. In these scenarios, models often need to identify informative local elements, suppress redundant tokens, and reason over structured relations among visual or multimodal components. A promising future direction is to integrate H$^3$Former with vision-language models, where hypergraph-based token-to-region aggregation may facilitate richer semantic interpretation and more flexible language-guided supervision.

\bibliographystyle{IEEEtran}
\bibliography{ref}

\begin{IEEEbiography}[{\includegraphics[width=1in,height=1.25in,clip,keepaspectratio]{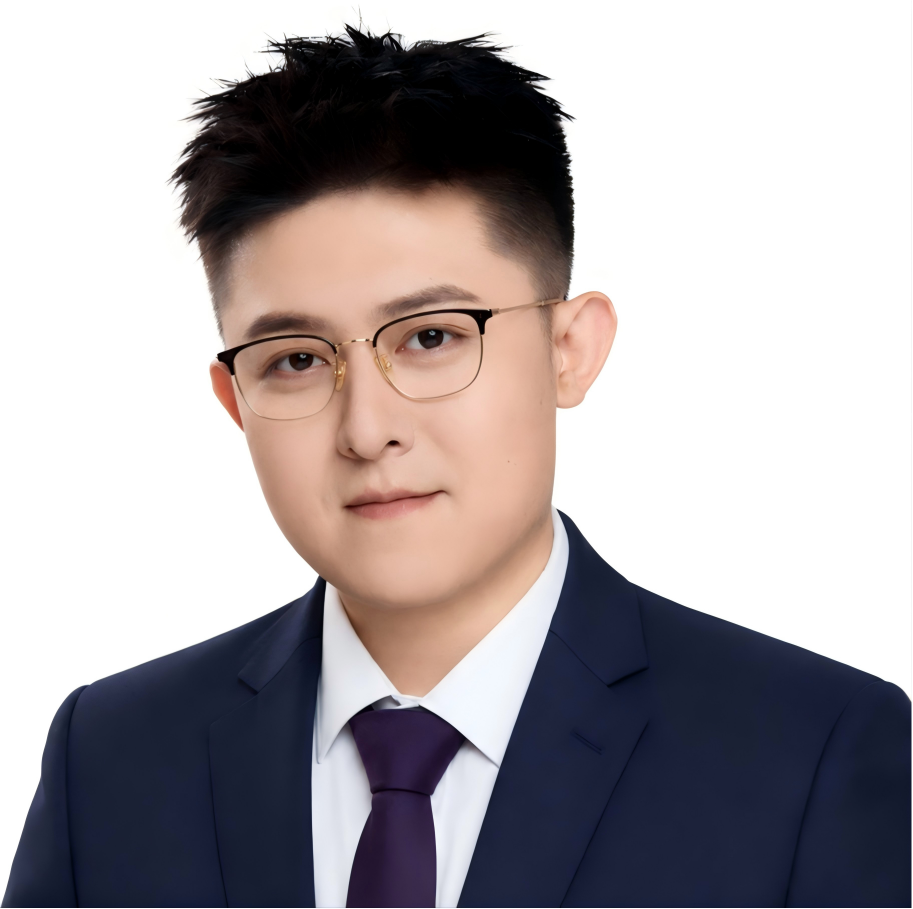}}]{Yongji Zhang} is currently working toward the Ph.D. degree in the College of Computer Science and Technology, Jilin University, China. He received the M.S. degree from the College of Computer Science and Technology, Jilin University, China, in 2023. His research interests include computer vision and machine learning. 
\end{IEEEbiography}

\begin{IEEEbiography}[{\includegraphics[width=1in,height=1.25in,clip,keepaspectratio]{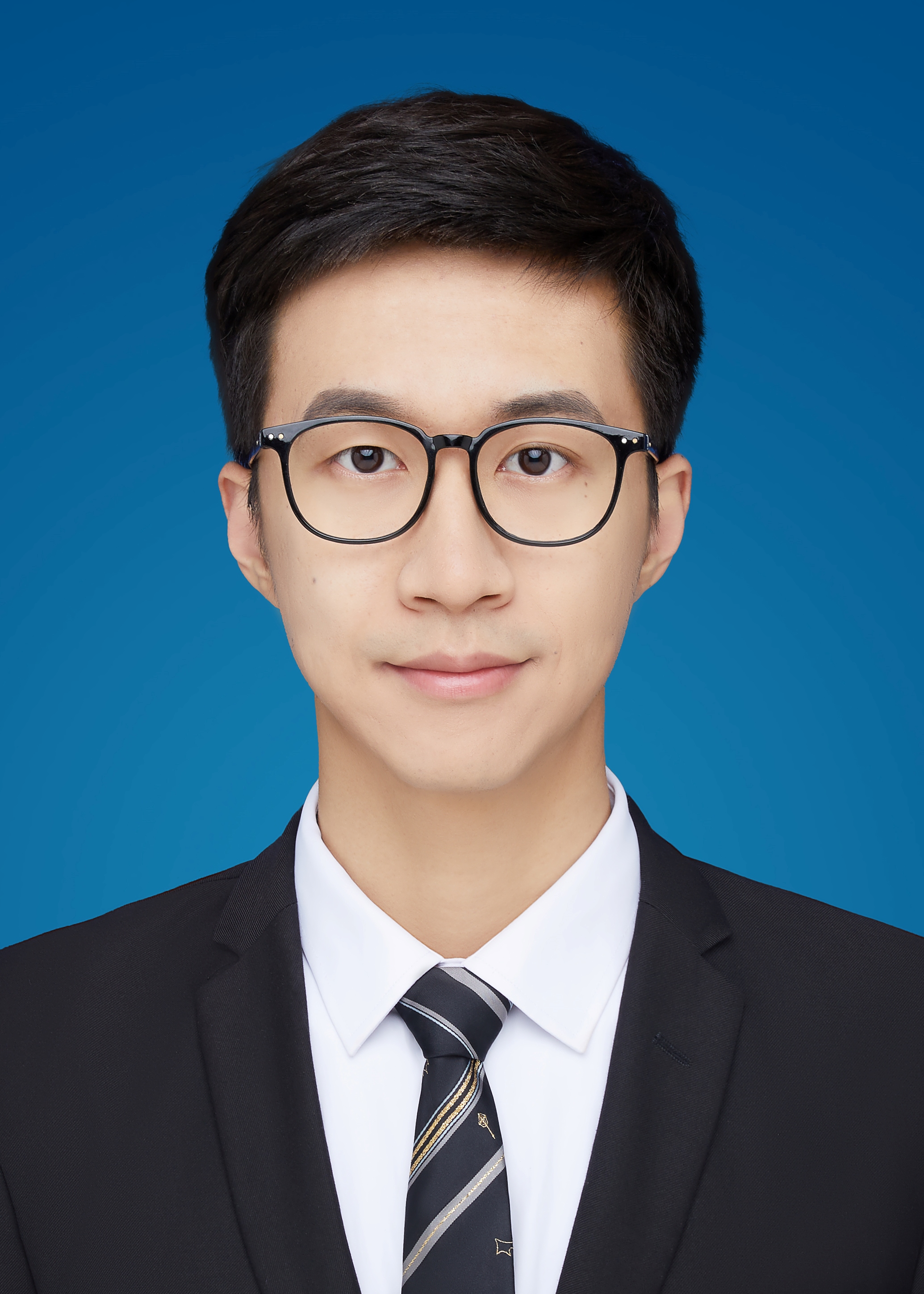}}]{Siqi Li} is currently a postdoctoral researcher at the School of Software, Tsinghua University, Beijing, China. He received the Ph.D. degree from Tsinghua University in 2024. His research interests include computer vision and machine learning. 
\end{IEEEbiography}

\begin{IEEEbiography}[{\includegraphics[width=1in,height=1.25in,clip,keepaspectratio]{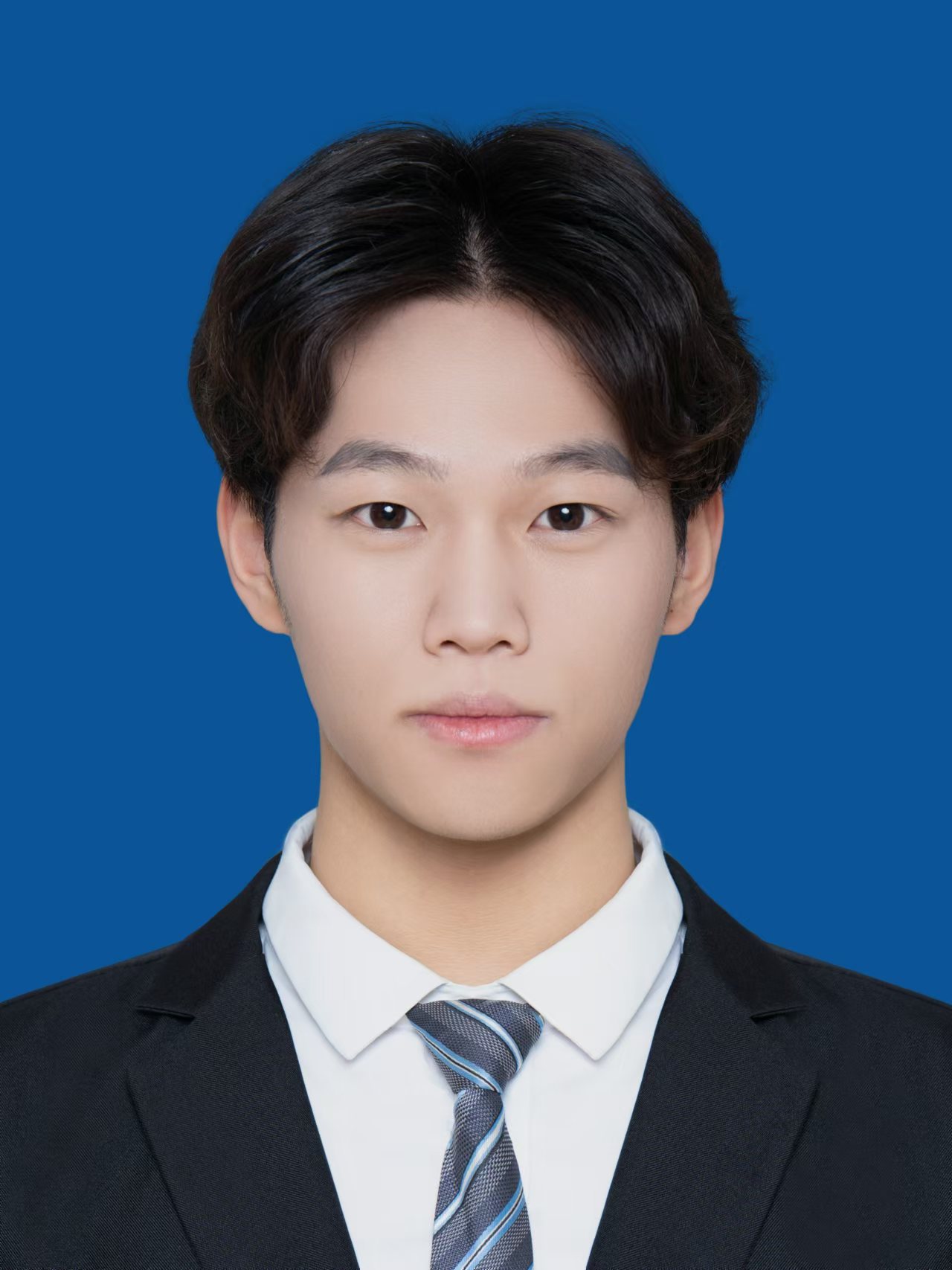}}]{Kuiyang Huang} received the B.S. degree from the Software College, Qingdao Agricultural University, China, in 2024. He is currently pursuing the M.S. degree with the School of Software, Jilin University, China. His research interests include computer vision and machine learning.
\end{IEEEbiography}

\begin{IEEEbiography}[{\includegraphics[width=1in,height=1.25in,clip,keepaspectratio]{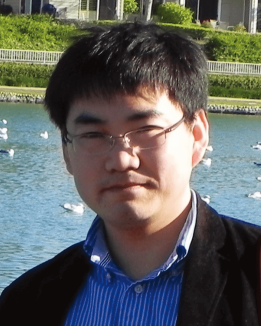}}]{Yue Gao} (Senior Member, IEEE) is an associate professor with the School of Software, Tsinghua University. He received the B.S. degree from the Harbin Institute of Technology, Harbin, China, and the M.E. and Ph.D. degrees from Tsinghua University, Beijing, China.
\end{IEEEbiography}

\begin{IEEEbiography}[{\includegraphics[width=1in,height=1.25in,clip,keepaspectratio]{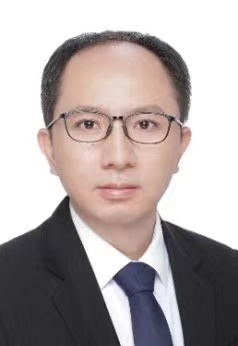}}]{Yu Jiang} (Member, IEEE) is a professor with the College of Computer Science and Technology, Jilin University, China. He received his M.S. and Ph.D. degrees from the College of Computer Science and Technology, Jilin University, China, in 2005 and 2011, respectively. His research fields include artificial intelligence and machine vision.
\end{IEEEbiography}
\end{document}